\documentclass{article} 
\usepackage{iclr2025_conference,times}
\pdfoutput=1

\usepackage{amsmath,amsfonts,bm}

\def\eqref#1{equation~\ref{#1}}

\def\1{\bm{1}}

\DeclareMathAlphabet{\mathsfit}{\encodingdefault}{\sfdefault}{m}{sl}
\SetMathAlphabet{\mathsfit}{bold}{\encodingdefault}{\sfdefault}{bx}{n}

\usepackage{hyperref}
\usepackage{url}
\usepackage{enumitem}
\usepackage{booktabs}
\usepackage{colortbl}
\usepackage{multirow}
\usepackage{xcolor}
\usepackage{graphicx}
\usepackage{subcaption}
\usepackage[many]{tcolorbox}   
\usepackage{epigraph} 
\definecolor{myblue}{RGB}{0, 0, 255}
\definecolor{mygray}{gray}{0.9}
\definecolor{goodblue}{RGB}{0,113,188}
\hypersetup{colorlinks=true,breaklinks,linkcolor=red,citecolor=goodblue}
\usepackage{xspace}
\usepackage{wrapfig}
\input{my_packages}

\title{PersonaMath: Boosting Mathematical Reasoning via Persona-Driven Data Augmentation}

\author{Jing Luo\textsuperscript{1}\thanks{Equal contribution \quad $\dag$ Corresponding author}
\authorskip Longze Chen\textsuperscript{1,2}$^*$
\authorskip Run Luo\textsuperscript{1,2}
\authorskip Liang Zhu\textsuperscript{1,2}
\authorskip Chang Ao\textsuperscript{1,2} 
\authorskip Jiaming Li\textsuperscript{1,2}\\
\textbf{Yukun Chen\textsuperscript{1,2}}
\textbf{\authorskip Xin Cheng\textsuperscript{1,4}}
\textbf{\authorskip Wen Yang\textsuperscript{5}}
\textbf{\authorskip Jiayuan Su\textsuperscript{6}}
\textbf{\authorskip Chengming Li\textsuperscript{7}}\\
\textbf{Ahmadreza Argha\textsuperscript{8}}
\textbf{\authorskip Hamid Alinejad-Rokny\textsuperscript{8}}
\textbf{\authorskip Shiwen Ni\textsuperscript{1}$^\dag$}
\textbf{\authorskip Min Yang\textsuperscript{1,3}$^\dag$} \\[2mm]
\textsuperscript{1}Shenzhen Key Laboratory for High Performance Data Mining, \\
Shenzhen Institutes of Advanced Technology, Chinese Academy of Sciences \\
\textsuperscript{2}University of Chinese Academy of Sciences \hspace{4mm} 
\textsuperscript{3}Shenzhen University of Advanced Technology \\
\textsuperscript{4}South China University of Technology \hspace{4mm} 
\textsuperscript{5}Institute of Automation, Chinese Academy of Sciences\\
\textsuperscript{6}Zhejiang University \hspace{4mm}
\textsuperscript{7}Shenzhen MSU-BIT University \hspace{4mm}
\textsuperscript{8}UNSW Sydney\\
\texttt{luojing020713@gmail.com}\\
\texttt{\{r.luo, lz.chen2, min.yang\}@siat.ac.cn} 
}

\iclrfinalcopy 
\begin{document}

\maketitle

\begin{abstract}
While closed-source Large Language Models (LLMs) demonstrate strong mathematical problem-solving abilities, open-source models still face challenges with such tasks. To bridge this gap, we propose a data augmentation approach and introduce PersonaMathQA, a dataset derived from MATH and GSM8K, on which we train the PersonaMath models. Our approach consists of two stages: the first stage focuses on learning from Persona Diversification, and the second stage emphasizes  learning from Reflection. In the first stage, we regenerate detailed chain-of-thought (CoT) solutions as instructions using a closed-source LLM and introduce a persona-driven data augmentation technique. This technique innovatively classifies personas based on occupations, significantly enhancing the dataset's diversity and quality. In the second stage, we incorporate reflection to fully leverage more challenging and valuable questions. Evaluation of our PersonaMath models on MATH and GSM8K reveals that the PersonaMath-7B model (based on Qwen2.5-7B) achieves an accuracy of 61.2\% on MATH and 87.8\% on GSM8K, surpassing all baseline methods and achieving state-of-the-art performance. Notably, our dataset contains only 128.9K data points—merely 32.6\% of MetaMathQA and 49.5\% of MathInstruct—yet our model outperforms these baselines, demonstrating the high quality and diversity of our dataset, which enables more efficient model training. 
We open-source the PersonaMathQA dataset, PersonaMath models, and our code for public usage.
\end{abstract}
\begin{figure*}[ht]
\centering
\includegraphics[width=0.9\linewidth]{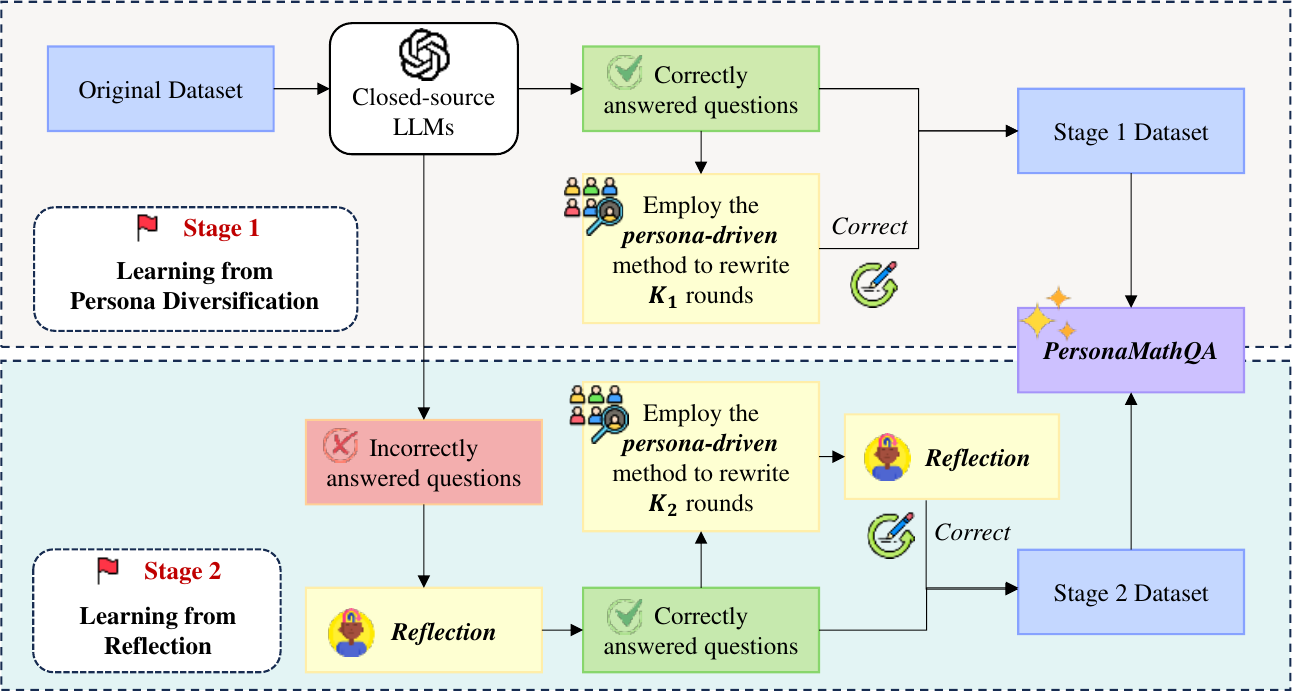}
\caption{The framework of our data augmentation method. The method consists of two stages: Stage 1 (top) and Stage 2 (bottom). Stage 1 focuses on using closed-source LLMs to automatically generate detailed CoT solutions and apply our persona-driven rewriting method to rephrase the questions.  Stage 2 focuses on reflection. The data from both stages are then combined to form our PersonaMathQA dataset.}
\label{framework}
\end{figure*}

\begin{figure*}[ht]
\centering
\includegraphics[width=0.45\linewidth]{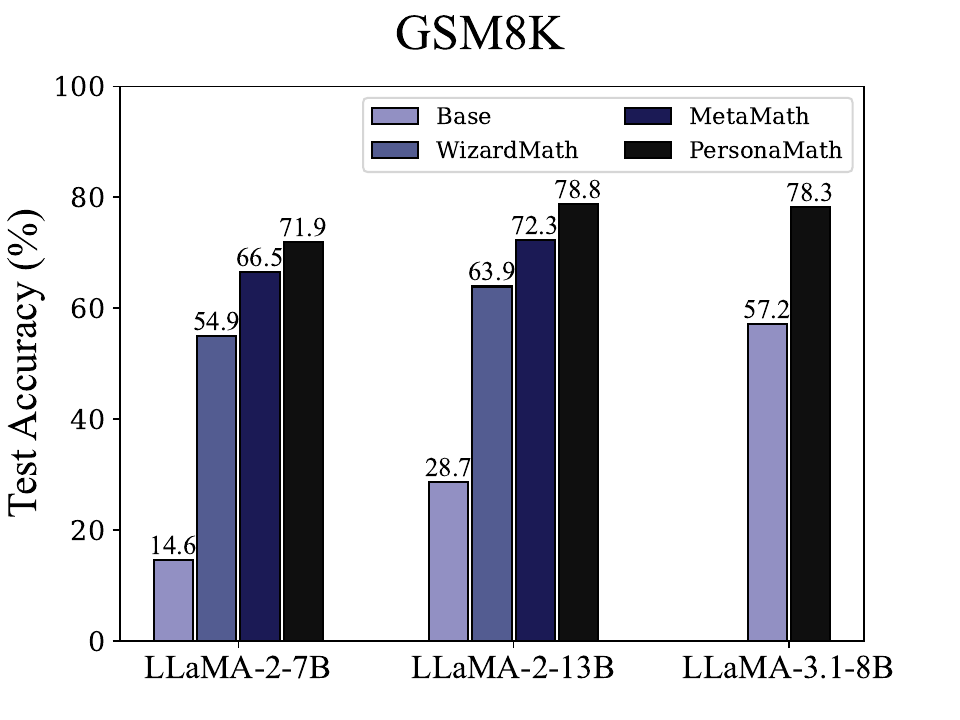} \hfill
\includegraphics[width=0.45\linewidth]{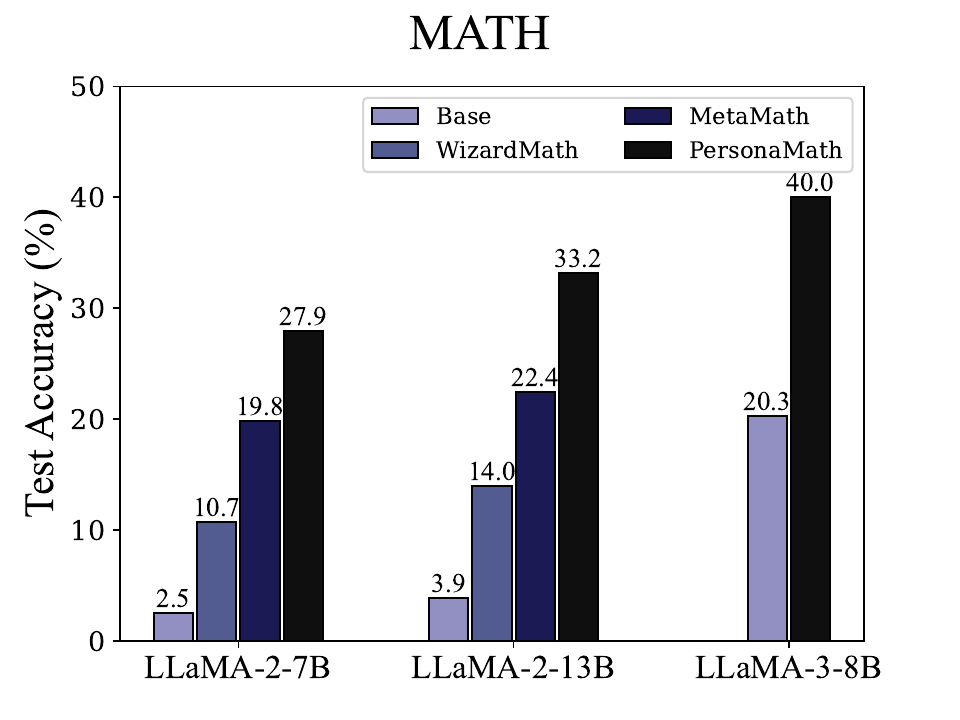}
\caption{The superior performance of our PersonaMath models in comparison to other models. Among all models of the same size, our model achieves the highest test accuracy, demonstrating state-of-the-art performance.}
\end{figure*}

\section{Introduction}
\setlength{\epigraphwidth}{0.95\columnwidth}
\renewcommand{\epigraphflush}{center}
\renewcommand{\textflush}{flushepinormal}
\renewcommand{\epigraphsize}{\footnotesize}
\epigraph{\textcolor{black}{\textit{``There are a thousand Hamlets in a thousand people's eyes''}}}
{\textcolor{black}{\textit{Shakespeare}}}

Large language models (LLMs) have rapidly advanced in recent years, showcasing impressive abilities across various Natural Language Processing (NLP) tasks, including open-domain conversation~\citep{NEURIPS2020_1457c0d6, ouyang2022traininglanguagemodelsfollow}, coding~\citep{chen2021evaluatinglargelanguagemodels, rozière2024codellamaopenfoundation, li2023starcodersourceyou}, and math~\citep{luo2023wizardmathempoweringmathematicalreasoning, yu2024metamathbootstrapmathematicalquestions, imani2023mathpromptermathematicalreasoningusing}. Among these tasks, solving math problems stands out as particularly challenging due to its complexity and the need for multi-step reasoning to reach a solution. While some closed-source models, such as GPT-4o~\citep{gpt4o}, Claude 3.5 Sonnet~\citep{Anthropic}, and Gemini 1.5 Pro~\citep{reid2024gemini}, have demonstrated strong math-solving capabilities, current open-source models (e.g., LLaMA~\citep{touvron2023llama, dubey2024llama}) continue to struggle in this domain. Therefore, enhancing the math problem-solving abilities of open-source models remains a prominent desiderata.
%

A widely adopted and effective approach for improving the math-solving capabilities of open-source models is fine-tuning, thanks to the accessibility of their weights~\citep{yuan2023scalingrelationshiplearningmathematical, yue2023mammothbuildingmathgeneralist, luo2023wizardmathempoweringmathematicalreasoning, yu2024metamathbootstrapmathematicalquestions}. However, directly fine-tuning on the original dataset yields only minimal improvements~\citep{touvron2023llama}, as solving complex math problems requires multiple reasoning steps, which the original dataset fails to adequately capture. Furthermore, the limited number of samples in the original dataset is often insufficient for models to fully learn how to solve mathematical problems. To address this, recent studies have demonstrated promising results by bootstrapping the original dataset~\citep{yu2024metamathbootstrapmathematicalquestions}, generating more detailed Chain of Thought (CoT) solutions~\citep{wei2023chainofthoughtpromptingelicitsreasoning}, and training models on the enhanced dataset. However, directly bootstrapping from the original problems can reduce the diversity of the enhanced dataset, requiring a large amount of data for effective training, which is both resource-intensive and~inefficient.

In this paper, we propose a data augmentation method aimed at enabling models to learn more efficiently and effectively from enhanced datasets. The key to our approach lies in improving both the quality and diversity of the data. To achieve this, we introduce a persona-driven approach and create PersonaMathQA, a dataset characterized by high efficiency and diversity.
Recently, Role-Playing LLMs that incorporate personas have gained significant attention, as leveraging personas allows LLMs to adapt to persona-specific scenarios and generate unique, diverse answers from various perspectives. Moreover, recent research has further highlighted the potential of Role-Playing LLMs. For instance, \citet{chan2024scalingsyntheticdatacreation} automatically created Persona Hub, a collection of 1 billion diverse personas, with 200,000 personas released publicly. Building on this resource, the authors generated numerous persona-specific queries to enhance model training. However, their methodology did not fully exploit the rich informational potential inherent in individual personas. To address this limitation and further enhance the diversity of generated data, we propose a novel approach that identifies the occupational characteristics of different roles and categorizes them into 11 distinct groups based on the International Standard Classification of Occupations (ISCO-08). Leveraging this classification, our persona-driven data rewriting method produces more nuanced and diverse data, significantly improving training effectiveness and efficiency.

Specifically, our method is divided into two stages. 
In the first stage, we focus on empowering open-source models to \textbf{Learn from Persona Diversification} by proposing a persona-driven method. We begin by using a powerful closed-source LLM to generate detailed CoT solutions for the math problems within the original dataset. After identifying the correctly answered questions, we apply the persona-driven method to prompt the closed-source LLM to rewrite these questions using specific and diverse personas. 
We then regenerate the CoT solutions and incorporate both the correctly answered questions and their original questions in our dataset. 
In the second stage, we shift our focus to enabling open-source models to \textbf{Learn from Reflection}. For questions that the closed-source LLM answered incorrectly in the first stage, we allow the LLM to reflect on its mistakes and attempt to regenerate the correct answers. For those questions answered correctly after reflection, which are inherently more challenging than those answered correctly in the first stage, we aim to increase their weight in the final dataset. This strategy facilitates more effective learning of these difficult questions by open-source LLMs during training.

Our main contributions are summarized as follows:
\begin{itemize}[topsep=0pt, before=\setlength{\itemsep}{0pt}, parsep=0pt, leftmargin=*]
\item We propose a novel persona-driven data augmentation method, resulting in the creation of a high-quality and efficient dataset, PersonaMathQA.
\item 
Using our PersonaMathQA dataset, we trained LLaMA and Qwen2.5, resulting in the PersonaMath models. 
We evaluate the math problem-solving capabilities of PersonaMath models, trained on our PersonaMathQA dataset, on two in-domain and two out-of-domain math benchmarks. Despite PersonaMathQA being the smallest in size compared to previous state-of-the-art (SOTA) datasets, the PersonaMath-7B model surpasses them on both benchmarks, setting a new SOTA performance.
\item We introduce a method to classify personas based on their occupations. By Leveraging this classification, we generate more diverse and nuanced data, providing valuable insights for future research to further explore and utilize  the rich information embedded within personas.
\end{itemize}

\section{Related Work}

\paragraph{Large Language Models for Mathematical Reasoning.} Among various NLP tasks, solving math problems has consistently been regarded as one of the most challenging. Current cutting-edge closed-source LLMs, such as o1~\citep{o1}, GPT-4o~\citep{gpt4o}, Claude 3.5 Sonnet~\citep{Anthropic}, and Gemini 1.5 Pro~\citep{reid2024gemini}, exhibit strong math problem-solving capabilities. Nonetheless, there remains substantial work to further enhance these capabilities through various strategies, including preprocessing math questions~\citep{an2023does}, utilizing more advanced prompts~\citep{ling-etal-2017-program, yao2023solvingmathwordproblem}, employing external tools~\citep{yamauchi2023lpmlllmpromptingmarkuplanguage, heyueya2023solvingmathwordproblems, chen-etal-2022-unigeo}, and enhancing overall interactions~\citep{wu2024mathchatconversetacklechallenging}. In contrast, open-source models with fewer parameters still struggle with solving math problems. Consequently, numerous studies, including our own, aim to bridge this gap, enabling open-source models to achieve robust math problem-solving capabilities after fine-tuning. Past approaches to enhance fine-tuning have primarily focused on generating intermediate steps for answers~\citep{nye2021workscratchpadsintermediatecomputation, zhang2023interpretablemathwordproblem, yang2023gptsolvemathematicalproblems, lewkowycz2022solvingquantitativereasoningproblems}, fine-tuning across multiple datasets~\citep{mishra-etal-2022-lila, yue2023mammothbuildingmathgeneralist}, employing teacher-student knowledge distillation~\citep{imani2023mathpromptermathematicalreasoningusing}, and learning from enhanced datasets~\citep{imani2023mathpromptermathematicalreasoningusing, raiyan-etal-2023-math, yu2024metamathbootstrapmathematicalquestions}, as explored in our paper.

\paragraph{Role-Playing Large Language Models.} Recently, Role-Playing LLMs that integrate personas have gained significant attention. 
Specifically, Role-Playing LLMs involve directly incorporating personas into the prompts of LLMs, enabling them to generate unique, role-specific content within a designated context. 
Although simple to implement by modifying prompts, this approach has yielded impressive results. For instance, \citet{dong2024selfcollaborationcodegenerationchatgpt} employed multiple LLM agents, assigning each to a specialized expert role, which enabled them to collaboratively tackle complex code generation tasks. Similarly, ChatDev~\citep{qian2024chatdevcommunicativeagentssoftware} segmented the software development process into four stages—designing, coding, testing, and documenting—and proposed a \textit{Chat Chain} to facilitate communication among agents responsible for each stage, ultimately enabling cooperative program development. Role-Playing LLMs are also frequently used in gaming contexts. For instance, \citet{wang2023voyageropenendedembodiedagent} employed an LLM to assume the role of a general assistant, which continuously explored the Minecraft game world to acquire skills and survive longer. 
In the medical domain, \citet{tang-etal-2024-medagents} simulates real-life scenarios by dividing the diagnostic reasoning process into five stages: expert gathering, analysis proposition, report summarization, collaborative consultation, and decision-making. Multiple LLM agents are then assigned different roles to collaborate on the diagnostic process.
\section{Method}

In this section, we provide a detailed description of how we constructed  PersonaMathQA. An overview of our approach is illustrated in Figure~ref{framework}. Our method, which enhances data through a persona-driven approach, is divided into two stages. Using this approach, we created the PersonaMathQA dataset by augmenting the MATH and GSM8K datasets and subsequently fine-tuned the PersonaMath models on the enhanced dataset.

\subsection{Stage 1: Learning from Persona Diversification}

\paragraph{Limitations of the Original Dataset.} Fine-tuning open-source models can significantly enhance their performance across various tasks, but improving their math problem-solving capabilities through fine-tuning alone presents  unique challenges. One reason is that \textit{\textbf{math problems are inherently complex}} and often require intricate multi-step reasoning to arrive at the correct answer~\citep{ahn-etal-2024-large}, \textit{\textbf{with errors at any step potentially leading to incorrect results.}} Another challenge stems from the limitations of current training datasets. Effective math problem-solving requires detailed, step-by-step reasoning; however \textit{\textbf{existing datasets often lack such comprehensive solutions, providing only final answers or insufficiently detailed explanations.}} Consequently, models trained on these datasets struggle to develop necessary reasoning skills, eading to only minimal improvements from fine-tuning~\citep{touvron2023llama}. While manually labeling detailed solutions is time-consuming and labor-intensive, we address this issue by leveraging a closed-source LLM with advanced mathematical problem-solving capabilities to automatically generate detailed CoT solutions. This process enhances the training dataset with comprehensive, step-by-step reasoning. The prompt we used can be found in Appendix~\ref{sub: Inference}.

\paragraph{Persona-Driven Data Augmentation Method.} After obtaining the CoT solutions automatically generated by the LLM, we first filter out the correct answers and add them to our PersonaMathQA dataset. However, this initial training data is insufficient for effective model improvement. To create a high-quality, enhanced dataset without excessive time and resource expenditure, we employ a data augmentation method in which the LLM automatically rewrites questions to generate new ones. Previous studies have shown that directly rewriting questions can be inefficient~\citep{yu2024metamathbootstrapmathematicalquestions}, as it requires a large volume of data to significantly boost model performance. This inefficiency stems from the homogeneous nature of the rewritten questions, which often lack diversity. To address this challenge and enhance the dataset more effectively, we propose a novel persona-driven method. Inspired by the observation that introducing persona-related scenarios to the LLM can prompt it to generate unique and persona-specific content, we hypothesize that with a sufficiently diverse set of personas, we can generate a large quantity of varied content that enhances both the quality and diversity of the dataset. 

A recent study~\citep{chan2024scalingsyntheticdatacreation} that released 200,000 diverse personas created from extensive web data provides a solid foundation for our approach. These personas encompass various nationalities, races, religions, occupations, age groups, and more, and were used to generate a wide variety of questions. However, they did not segment the personas during the generation process, which led to a lack of diversity in the generated output. This limitation arises because data generated using similar personas tend to exhibit high similarity. To overcome this challenge, we propose a \textit{\textbf{Persona-Driven Data Augmentation Method}}, which first segments the personas into distinct groups and then selects different persona types for data generation in each iteration. This approach significantly enhances the diversity of the generated data. 

Specifically, we observed that the personas in Persona Hub often include occupational characteristics, such as "A Political Analyst specialized in El Salvador's political landscape". This observation led us to the conclusion that personas can be systematically classified based on their occupations. Consequently, we categorized the personas into 11 distinct groups using the International Standard Classification of Occupations (ISCO-08). To automate this classification process, we employed closed-source LLM, and the prompt used for this task can be found in Appendix \ref{sub: Classification}. ISCO-08 divides occupations into 10 major groups, including Managers, Professionals, Technicians and Associate Professionals, among others. To accommodate personas whose occupations could not be clearly identified by the LLM, we introduced an additional group labeled \textit{Others}, ensuring comprehensive coverage of all personas in the dataset.

We then leverage these classified personas to guide the LLM in rewriting each question. In each round, we select one persona from each of the 11 occupational categories and instruct the LLM to rewrite the question based on the selected persona, resulting in 11 distinct rewrites per question. The prompt we used can be found in Appendix \ref{sub: Rewriting}. In Stage 1, we perform $K_1$ rounds of rewriting. Afterward, the LLM reasons through these rewritten questions and generates detailed CoT solutions. We select the correct solutions, along with their corresponding questions, to include in our PersonaMathQA dataset. Questions that are answered incorrectly after rewriting are discarded.

\subsection{Stage 2: Learning from Reflection}

\paragraph{Shifting Focus to Incorrectly Answered Questions.} In Stage 1, we first have the LLM reason through the original dataset and identify questions with correct answers. Questions answered incorrectly are set aside for further analysis. For this analysis, we use the MATH dataset, where each problem is classified into difficulty levels ranging from `1' to `5'. We calculated the average difficulty levels of correctly and incorrectly answered questions during the reasoning stage. The results indicate that the average difficulty level of correctly answered questions is 3.22, while incorrectly answered questions have an average level of 4.28. This suggests that the incorrectly answered questions are inherently more challenging, making them harder for the LLM to solve. However, this also makes them more valuable, as learning to solve these complex problems can significantly enhance the model's step-by-step reasoning ability.

\paragraph{Reflecting on Errors and Regenerating Solutions.} To harness the potential value of the incorrectly answered questions, we prompt the LLM to reflect on its errors and attempt to provide correct answers. This approach draws inspiration from previous research demonstrating that LLMs can self-reflect and self-correct~\citep{shinn2023reflexionlanguageagentsverbal,li-etal-2023-making}. Specifically, for the questions that the LLM answered incorrectly in Stage 1, we present the incorrect solution to the LLM, inform it of the error, and prompt it to reflect on its reasoning process before generating a corrected, detailed CoT solution. The prompt we used can be found in Appendix \ref{sub: Reflection}, where the explanation is updated with the incorrect solution.

In this step, the LLM’s response is divided into two parts: ``Review of Incorrect Explanation'' and ``Corrected Explanation''. We use only the ``Corrected Explanation'' as the regenerated solution from the LLM. However, it is important to note that while the ``Review of Incorrect Explanation" is not used directly, it plays an indispensable role in the reflection process. This part allows the LLM to summarize its mistakes, which is essential for generating the subsequent "Corrected Explanation". Correctly answered questions are then added to our PersonaMathQA dataset, while questions that remain unanswered correctly by the LLM are discarded. 
It is worth noting that previous studies have also explored generating the correct solution process by directly providing the LLM with the correct answer~\citep{zhang2024sciglmtrainingscientificlanguage}. However, this approach carries risks due to LLM hallucinations, where the LLM may generate an incorrect solution process while providing the correct final answer. Such data can be misleading and detrimental. Therefore, we discard questions that cannot be answered correctly.

For the questions answered correctly in Stage 2, we apply the same persona-driven rewriting method as in Stage 1. The key difference in Stage 2 is that we perform $K_2$ rounds of rewriting, where $K_2$ is greater than $K_1$, the number of rewriting rounds in Stage 1. As discussed earlier, Stage 2 problems are more challenging, and the performance gains from training on these difficult problems are more substantial. Therefore, increasing the number of rewrites in Stage 2 helps enhance model performance more effectively after training. We then apply our reflection framework to regenerate the solutions, selecting only those with  correct answers. For computational efficiency, we use the incorrect solution from the original question as the incorrect solution for the rewritten questions.

\section{Experiments}

\begin{table*}[!t]
\begin{center}
\scalebox{0.85}{
\begin{tabular}{lcccccc}
\toprule[1pt]
\multirow{2}{*}{Model}&  \multirow{2}{*}{Params} & \multirow{2}{*}{Base} & \multicolumn{2}{c}{In-Domain} & \multicolumn{2}{c}{Out-of-Domain}\\
 & & & GSM8K & MATH & College & DM\\ 
\hline
\multicolumn{5}{c}{\textit{closed-source models}}\\
\hline
GPT-4~\citep{openai2024gpt4technicalreport} & - & - & 92.0 & 42.5 & - & -\\
o1~\citep{o1} & - & - & - & 94.8 & - & -\\
Claude 3.5 Sonnet~\citep{Anthropic} & - & - & 96.4 & 71.1 & - & -\\
Gemini 1.5 Pro~\citep{reid2024gemini} & - & - & 90.8 & 67.7 & - & -\\
\hline
\multicolumn{5}{c}{\textit{open-source models (6-9B)}}\\
\hline
LLaMA-2-7B~\citep{touvron2023llama} & 7B & - & 14.6 & 2.5 & 2.3 & -\\
LLaMA-3-8B~\citep{dubey2024llama} & 8B & - & 57.2 & 20.3 & - & -\\
GLM-4-9B~\citep{glm2024chatglm} & 9B & - & 84.0 & 30.4 & - & -\\
Qwen2.5-7B~\citep{qwen2.5} & 7B & - & 85.4 & 49.8 & - & - \\
DeepSeek-V2~\citep{deepseekai2024deepseekv2strongeconomicalefficient} &  16B & - & 41.1 & 17.1 & - & -\\
SFT~\citep{touvron2023llama} & 7B & LLaMA-2-7B & 41.6 & - & - & - \\
RFT~\citep{yuan2023scalingrelationshiplearningmathematical} & 7B & LLaMA-2-7B & 50.3 & - & - & -\\
WizardMath~\citep{luo2023wizardmathempoweringmathematicalreasoning} & 7B & LLaMA-2-7B & 54.9 & 10.7 & 6.8 & - \\
MAmmoTH(CoT)~\citep{yue2023mammothbuildingmathgeneralist} & 7B & LLaMA-2-7B & 50.5 & 10.4 & 6.2 & -\\
MetaMath~\citep{yu2024metamathbootstrapmathematicalquestions} & 7B & LLaMA-2-7B & 66.5 & 19.8 & 9.4 & -\\
Xwin-Math~\citep{li2024common7blanguagemodels} & 7B & LLaMA-2-7B & 84.9 & 45.5 & 27.6 & 43.0\\
DART-Math~\citep{tong2024dartmathdifficultyawarerejectiontuning} & 8B & LLaMA-3-8B & 81.1 & 46.6 & 28.8 & 48.0\\
\rowcolor{gray!25}\textbf{PersonaMath} & 7B & LLaMA-2-7B & \textbf{71.9} & \textbf{28.4} & \textbf{15.5} & \textbf{27.9} \\
\rowcolor{gray!25}\textbf{PersonaMath} & 8B & LLaMA-3-8B & \textbf{78.3} & \textbf{40.7} & \textbf{23.3} & \textbf{40.0} \\
\rowcolor{gray!25}\textbf{PersonaMath} & 7B & Qwen2.5-7B & \textbf{87.8} & \textbf{61.2} & \textbf{44.6} & \textbf{71.5} \\
\hline
\multicolumn{5}{c}{\textit{open-source models (more than 10B)}}\\
\hline
LLaMA-2-13B~\citep{touvron2023llama} & 13B & - & 28.7 & 3.9 & 1.2 & -\\
LLaMA-2-70B~\citep{touvron2023llama} & 70B & - & 56.8 & 13.5 & - & -\\
LLaMA-3-70B~\citep{dubey2024llama} & 70B & - & 83.7 & 41.4 & - & -\\
Qwen2.5-72B~\citep{qwen2.5} & 72B & - & 91.5 & 62.1 & - & -\\
DeepSeek-V2~\citep{deepseekai2024deepseekv2strongeconomicalefficient} &  236B & - & 79.2 & 43.6 & - & -\\
SFT~\citep{touvron2023llama} & 13B & LLaMA-2-13B & 50.0 & - & - & -\\
RFT~\citep{yuan2023scalingrelationshiplearningmathematical} & 13B & LLaMA-2-13B & 55.4 & - & - & -\\
WizardMath~\citep{luo2023wizardmathempoweringmathematicalreasoning} & 13B & LLaMA-2-13B & 63.9 & 14.0 & 7.8 & - \\
MAmmoTH(CoT)~\citep{yue2023mammothbuildingmathgeneralist} & 13B & LLaMA-2-13B & 56.3 & 12.9 & 6.5 & -\\
MetaMath~\citep{yu2024metamathbootstrapmathematicalquestions} & 13B & LLaMA-2-13B & 72.3 & 22.4 & 10.1 & -\\
\rowcolor{gray!25}\textbf{PersonaMath} & 13B & LLaMA-2-13B & \textbf{78.8} & \textbf{33.4} & \textbf{18.0} & \textbf{33.2}\\
\bottomrule
\end{tabular}}
\end{center}
\caption{Test accuracy on GSM8K and MATH datasets. ``PersonaMath" refers to our model. The primary baselines compared include WizardMath, MAmmoTH, and MetaMath. Our models demonstrate higher accuracy compared to the baselines when trained on the same underlying model}
\label{main result}
\end{table*}

\subsection{Training and Evaluation Setup}
We fine-tune open-source models using the PersonaMathQA dataset to develop the PersonaMath models.The training process follows prompt provided in~\citep{alpaca}, where we maximize the log-likelihood of the reasoning path given the question. Specifically, the loss function is expressed as $\mathcal{L}(\theta)=\Sigma_{(q,r)\in {\rm PersonaMathQA}}\log P(r|q;\theta)$. Here, $\theta$ represents the parameters of the open-source model, $q$ represents the question, and $r$ denotes the correct solution generated by the closed-source LLM. This formulation ensures that our method is easily adaptable for fine-tuning any open-source model. The training prompt, which replaces the ``\textit{instruction}" which replaces the ``\textit{instruction}" with questions from the PersonaMathQA dataset and follows the corresponding solutions after ``\textit{Response: }", is provided in Appendix~\ref{sub: Training}.
The CoT prompt used for evaluating the trained model is sourced from~\citep{alpaca}, as provided in Appendix~\ref{sub: Evaluation}. The ``\textit{instruction}" is replaced with questions from the test set.

\subsection{Experimental Setup}

\paragraph{Datasets.} We apply our data-enhancement method to two well-known math problem datasets, MATH~\citep{hendrycks2021measuringmathematicalproblemsolving} and GSM8K~\citep{cobbe2021trainingverifierssolvemath}, to create our PersonaMathQA dataset. GSM8K consists of problems designed by human writers, requiring between 2 and 8 steps to solve. Solutions involve a sequence of basic arithmetic operations (+, -, *, /) to arrive at the final answer. The GSM8K dataset includes approximately 7,500 training problems and about 1,000 test problems. In contrast, MATH is a more challenging dataset composed of problems from mathematics competitions, such as AMC 10, AMC 12, and AIME, covering a wide range of subjects and difficulty levels. The problems span seven subjects: Prealgebra, Algebra, Number Theory, Counting and Probability, Geometry, Intermediate Algebra, and Precalculus, and are classified into five difficulty levels, with higher numbers indicating greater complexity. MATH consists of 7,500 training problems and 5,000 test problems. 

To evaluate our model comprehensively, we conducted both in-domain and out-of-domain evaluations. For in-domain evaluation, we used the test sets from MATH and GSM8K. For out-of-domain evaluation, we followed the approach of DART-Math~\citep{tong2024dartmathdifficultyawarerejectiontuning} and included the following two additional test sets:
\begin{itemize}[topsep=0pt, before=\setlength{\itemsep}{0pt}, parsep=0pt, leftmargin=*]
\item \textbf{CollegeMath}~\citep{tang2024mathscalescalinginstructiontuning}: This test set contains 2,818 test samples sourced from 9 college mathematics textbooks. It covers seven critical mathematical disciplines, providing a robust evaluation of advanced mathematical reasoning.
\item \textbf{DeepMind-Mathematics}~\citep{saxton2019analysing}: This test set consists of 1,000 test examples based on a national school mathematics curriculum (up to age 16). It spans 8 different topics, offering a comprehensive range of mathematical concepts aligned with educational progression.
\end{itemize}

\paragraph{Models.} For classification, inference, rewriting, and reflection, we utilized the closed-source LLM GPT-4o-mini-2024-07-18~\citep{gpt4omini} with a temperature setting of 0.7. During the rewriting phase, \textit{\textbf{we configured the number of rewriting rounds as $K_1=1$ and $K_2=2$}}. To evaluate our approach, we fine-tuned the open-source models LLaMA-2-7B and LLaMA-2-13B~\citep{touvron2023llama} on our dataset. Additionally, to facilitate future comparisons, we also fine-tuned the current SOTA open-source model LLaMA-3-8B~\citep{dubey2024llama} and Qwen2.5-7B~\citep{qwen2.5}. For more training details, please refer to Appendix \ref{sec: exp}.

\paragraph{Baselines.} To evaluate our proposed method comprehensively, we establish a baseline that includes a range of closed-source models, open-source models, and SOTA methods for training open-source models to solve mathematical problems. The closed-source models include o1-preview~\citep{o1}, GPT-4o~\citep{gpt4o}, Claude 3.5 Sonnet~\citep{Anthropic}, and others. The open-source models encompass LLaMA-2, LLaMA-3, Qwen2.5, and others. Additionally, we compared our method against several prominent techniques, including WizardMath~\citep{luo2023wizardmathempoweringmathematicalreasoning}, MAmmoTH~\citep{yue2023mammothbuildingmathgeneralist}, MetaMath~\citep{yu2024metamathbootstrapmathematicalquestions}, Xwin-Math~\citep{li2024common7blanguagemodels}, and DART-Math~\citep{tong2024dartmathdifficultyawarerejectiontuning}. For these methods, we directly adopted the results reported in their respective papers.
Furthermore, we compared our method with several basic fine-tuning approaches:
\begin{itemize}[topsep=0pt, before=\setlength{\itemsep}{0pt}, parsep=0pt, leftmargin=*]
\item \textbf{Supervised Fine-Tuning (SFT)}: This method involves fine-tuning the models using the training sets from the original GSM8K or MATH datasets.
\item \textbf{Rejection Sampling Fine-Tuning (RFT)}~\citep{yuan2023scalingrelationshiplearningmathematical}: This approach generates and collects correct reasoning paths as augmented data for fine-tuning.
\end{itemize}

\subsection{Main Results}

Our PersonaMathQA dataset consists of 128.9K samples, with 74.7K generated based on GSM8K and 54.2K generated based on MATH. Additional details regarding the dataset composition can be found in Appendix \ref{sec: compos}. Table \ref{main result} presents the test accuracy of our method compared to a range of baselines. The results demonstrate that, compared to the pre-trained models, our trained model's mathematical problem-solving abilities have significantly improved, surpassing all open-source models of comparable size. Additionally, our method outperforms most baseline approaches, particularly our PersonaMath model based on Qwen2.5-7B, which achieves state-of-the-art results and outperforms all other models. Notably, among the baselines that also employ data enhancement techniques, the MathInstruct dataset used in MAmmoTH contains 260K samples, and the MetaMathQA dataset includes 395K samples. Despite our PersonaMathQA dataset having only 128.9K samples—49.5\% of MathInstruct and 32.6\% of MetaMathQA—our approach yields superior results. This indicates that our dataset is of higher quality, enabling the model to acquire more robust mathematical problem-solving skills with fewer samples. However, when trained on the same base model, our method still trails DART-Math. DART-Math employs a Rejection-Based Data Synthesis method and utilizes a dataset of 591K samples, whereas our dataset is only 21.8\% of its size. This indicates that expanding the PersonaMathQA dataset using our method could further enhance model performance, potentially closing the gap with DART-Math and achieving even greater improvements.

\subsection{Discussion on Dataset Diversity}

\begin{wrapfigure}{r}{0.5\linewidth}
\centering
\includegraphics[width=1\linewidth]{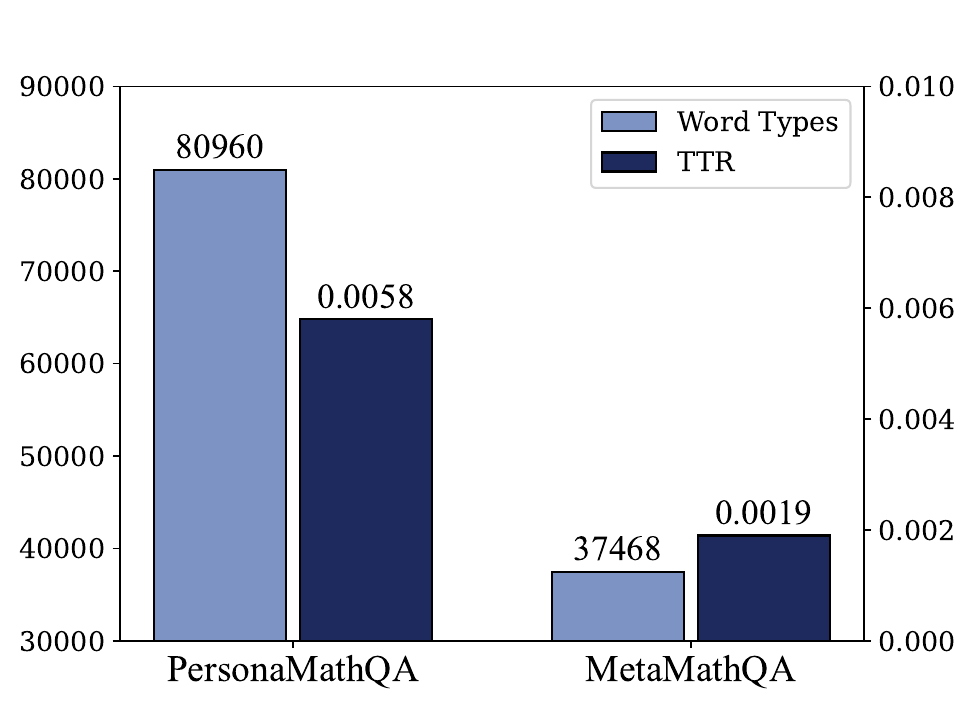}
\caption{Comparison of Word Types and TTR between our PersonaMathQA dataset and MetaMathQA. PersonaMathQA significantly surpasses MetaMathQA in both metrics, demonstrating its superior diversity and quality.}
\label{diversity}

\end{wrapfigure}

One notable aspect of our main results is that our method not only surpasses the baseline method MetaMath but also achieves this with a significantly smaller dataset, containing only 32.6\% of their samples. To illustrate this, we conducted two experiments to assess the diversity of questions in our dataset compared to the baseline datasets. In the first experiment, we split the questions into individual words, counted the Word Types (the number of unique, non-repeating words), and calculated the Type Token Ratio (TTR), which is the ratio of Word Types to the total number of words, for both our dataset and the MetaMathQA dataset. Higher values for Word Types and TTR indicate greater diversity. Our experimental results, shown in Figure \ref{diversity}, reveal that the PersonaMathQA dataset significantly outperforms MetaMathQA in both Word Types and TTR. 

\begin{figure}[t]
\begin{center}
\centering
\includegraphics[width=0.9\linewidth]{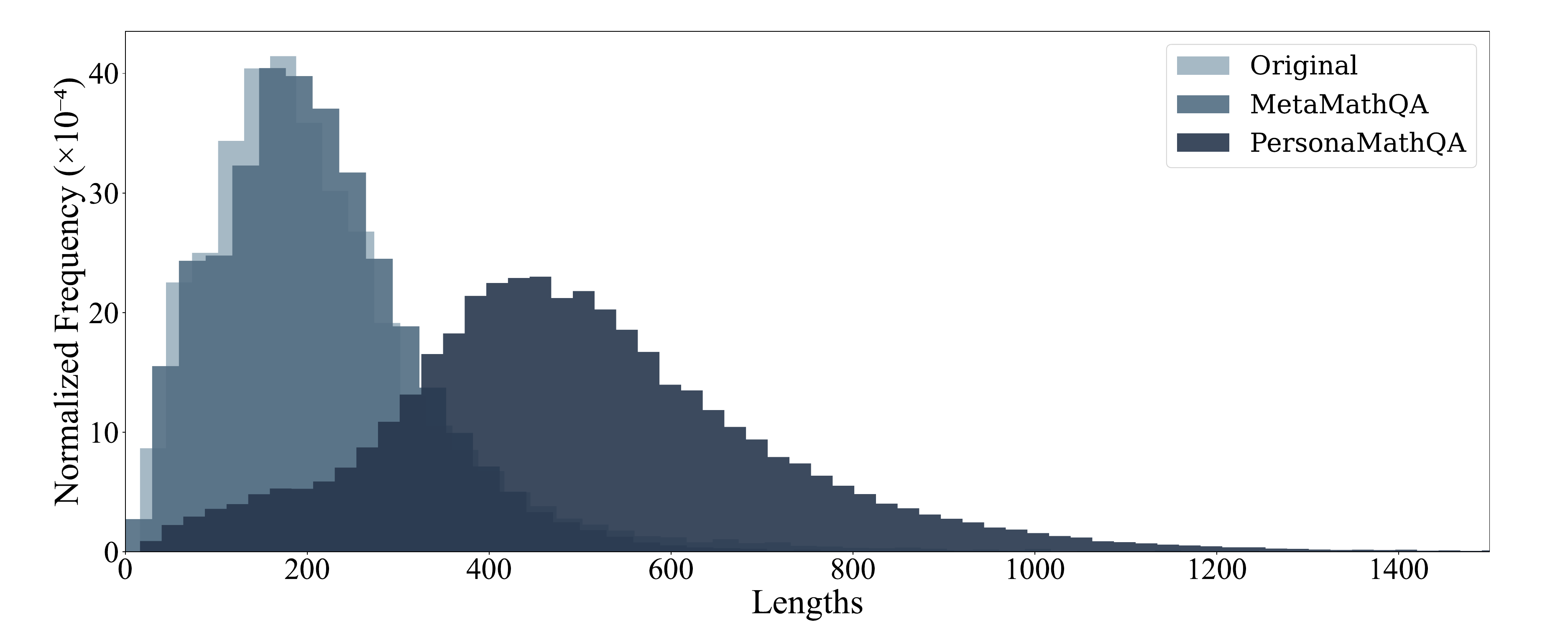}
\end{center}
\caption{Comparison of the distribution of question lengths between our dataset and the two baseline datasets, where ``Original" refers to the sum of the MATH and GSM8K datasets. The result shows that the distribution of question lengths in our dataset is more uniform and broader than in the two baseline datasets, indicating superior diversity.}
\label{frequency}

\end{figure}

In the second experiment, we calculated the distribution of question lengths across three datasets. The results, shown in Figure \ref{frequency}, where ``Original" refers to the sum of MATH and GSM8K datasets, and ``Normalized Frequency" refers to the frequency normalized so that the area under the histogram integrates to 1. As depicted in this figure, the question length distributions in the original dataset and MetaMathQA overlap significantly and are concentrated in the range of shorter questions, indicating a lack of uniformity in the distribution. In contrast, our dataset demonstrates a more uniform and broader distribution, featuring a higher proportion of longer questions and covering a wider variety of question lengths. Both experiments demonstrate that our dataset possesses markedly higher diversity. Consequently, our dataset is of superior quality and can achieve better results with fewer samples. This enhanced diversity stems from the persona-driven method employed, particularly the innovative classification of personas based on occupation. As illustrated by the example in Appendix \ref{sec: case}, even when starting from the same question, incorporating different personas introduces varied contexts for the LLM, leading to diverse and unique rewritten questions.

\section{Conclusion}

In this paper, we enhance the mathematical problem-solving capabilities of open-source LLMs by introducing a two-stage dataset augmentation method.
The first stage, Learning from Persona Diversification, involves generating detailed CoT solutions and applying a novel persona-driven approach for rewriting questions. 
The second stage, Learning from Reflection, leverages reflection to maximize more challenging and valuable questions. 
By applying this method to the MATH and GSM8K datasets, we developed PersonaMathQA, a high-quality, diverse dataset with 128.9K samples. Despite its smaller size than baseline methods, the PersonaMath models trained on PersonaMathQA have achieved SOTA performance across multiple test datasets. Our approach demonstrates the significant potential of persona-driven methods for dataset enhancement and provides valuable insights for future research in mathematical problem-solving.
\section*{Limitations}

When evaluating model performance, we are currently limited to assessing only whether the final answers derived through step-by-step reasoning are correct. This approach, however, fails to capture the nuanced quality of the model’s reasoning capabilities. For instance, consider two reasoning paths with incorrect results: one where the error occurs at the final step, and another where the error arises at the beginning. From a results-oriented perspective, both are incorrect; however, the former is less erroneous than the latter. To address this limitation, future work could integrate a Process Reward Model (PRM) to systematically distinguish between error origins, enabling a more granular and robust evaluation of reasoning processes.

\bibliographystyle{iclr2025_conference}
\bibliography{custom}

\begin{thebibliography}{51}
\providecommand{\natexlab}[1]{#1}
\providecommand{\url}[1]{\texttt{#1}}
\expandafter\ifx\csname urlstyle\endcsname\relax
  \providecommand{\doi}[1]{doi: #1}\else
  \providecommand{\doi}{doi: \begingroup \urlstyle{rm}\Url}\fi

\bibitem[Ahn et~al.(2024)Ahn, Verma, Lou, Liu, Zhang, and Yin]{ahn-etal-2024-large}
Janice Ahn, Rishu Verma, Renze Lou, Di~Liu, Rui Zhang, and Wenpeng Yin.
\newblock Large language models for mathematical reasoning: Progresses and challenges.
\newblock In Neele Falk, Sara Papi, and Mike Zhang (eds.), \emph{Proceedings of the 18th Conference of the European Chapter of the Association for Computational Linguistics: Student Research Workshop}, pp.\  225--237, St. Julian{'}s, Malta, March 2024. Association for Computational Linguistics.
\newblock URL \url{https://aclanthology.org/2024.eacl-srw.17}.

\bibitem[An et~al.(2023)An, Lee, and Gweon]{an2023does}
Jisu An, Junseok Lee, and Gahgene Gweon.
\newblock Does chatgpt comprehend the place value in numbers when solving math word problems?
\newblock In \emph{Human-AI Math Tutoring@ AIED}, pp.\  49--58, 2023.

\bibitem[Anthropic(2024)]{Anthropic}
Anthropic, 2024.
\newblock URL \url{https://www-cdn.anthropic.com/fed9cc193a14b84131812372d8d5857f8f304c52/Model_Card_Claude_3_Addendum.pdf}.

\bibitem[Brown et~al.(2020)Brown, Mann, Ryder, Subbiah, Kaplan, Dhariwal, Neelakantan, Shyam, Sastry, Askell, Agarwal, Herbert-Voss, Krueger, Henighan, Child, Ramesh, Ziegler, Wu, Winter, Hesse, Chen, Sigler, Litwin, Gray, Chess, Clark, Berner, McCandlish, Radford, Sutskever, and Amodei]{NEURIPS2020_1457c0d6}
Tom Brown, Benjamin Mann, Nick Ryder, Melanie Subbiah, Jared~D Kaplan, Prafulla Dhariwal, Arvind Neelakantan, Pranav Shyam, Girish Sastry, Amanda Askell, Sandhini Agarwal, Ariel Herbert-Voss, Gretchen Krueger, Tom Henighan, Rewon Child, Aditya Ramesh, Daniel Ziegler, Jeffrey Wu, Clemens Winter, Chris Hesse, Mark Chen, Eric Sigler, Mateusz Litwin, Scott Gray, Benjamin Chess, Jack Clark, Christopher Berner, Sam McCandlish, Alec Radford, Ilya Sutskever, and Dario Amodei.
\newblock Language models are few-shot learners.
\newblock In H.~Larochelle, M.~Ranzato, R.~Hadsell, M.F. Balcan, and H.~Lin (eds.), \emph{Advances in Neural Information Processing Systems}, volume~33, pp.\  1877--1901. Curran Associates, Inc., 2020.
\newblock URL \url{https://proceedings.neurips.cc/paper_files/paper/2020/file/1457c0d6bfcb4967418bfb8ac142f64a-Paper.pdf}.

\bibitem[Chan et~al.(2024)Chan, Wang, Yu, Mi, and Yu]{chan2024scalingsyntheticdatacreation}
Xin Chan, Xiaoyang Wang, Dian Yu, Haitao Mi, and Dong Yu.
\newblock Scaling synthetic data creation with 1,000,000,000 personas, 2024.
\newblock URL \url{https://arxiv.org/abs/2406.20094}.

\bibitem[Chen et~al.(2022)Chen, Li, Qin, Lu, Lin, Chen, and Liang]{chen-etal-2022-unigeo}
Jiaqi Chen, Tong Li, Jinghui Qin, Pan Lu, Liang Lin, Chongyu Chen, and Xiaodan Liang.
\newblock {U}ni{G}eo: Unifying geometry logical reasoning via reformulating mathematical expression.
\newblock In Yoav Goldberg, Zornitsa Kozareva, and Yue Zhang (eds.), \emph{Proceedings of the 2022 Conference on Empirical Methods in Natural Language Processing}, pp.\  3313--3323, Abu Dhabi, United Arab Emirates, December 2022. Association for Computational Linguistics.
\newblock \doi{10.18653/v1/2022.emnlp-main.218}.
\newblock URL \url{https://aclanthology.org/2022.emnlp-main.218}.

\bibitem[Chen et~al.(2021)Chen, Tworek, Jun, Yuan, de~Oliveira~Pinto, Kaplan, Edwards, Burda, Joseph, Brockman, Ray, Puri, Krueger, Petrov, Khlaaf, Sastry, Mishkin, Chan, Gray, Ryder, Pavlov, Power, Kaiser, Bavarian, Winter, Tillet, Such, Cummings, Plappert, Chantzis, Barnes, Herbert-Voss, Guss, Nichol, Paino, Tezak, Tang, Babuschkin, Balaji, Jain, Saunders, Hesse, Carr, Leike, Achiam, Misra, Morikawa, Radford, Knight, Brundage, Murati, Mayer, Welinder, McGrew, Amodei, McCandlish, Sutskever, and Zaremba]{chen2021evaluatinglargelanguagemodels}
Mark Chen, Jerry Tworek, Heewoo Jun, Qiming Yuan, Henrique~Ponde de~Oliveira~Pinto, Jared Kaplan, Harri Edwards, Yuri Burda, Nicholas Joseph, Greg Brockman, Alex Ray, Raul Puri, Gretchen Krueger, Michael Petrov, Heidy Khlaaf, Girish Sastry, Pamela Mishkin, Brooke Chan, Scott Gray, Nick Ryder, Mikhail Pavlov, Alethea Power, Lukasz Kaiser, Mohammad Bavarian, Clemens Winter, Philippe Tillet, Felipe~Petroski Such, Dave Cummings, Matthias Plappert, Fotios Chantzis, Elizabeth Barnes, Ariel Herbert-Voss, William~Hebgen Guss, Alex Nichol, Alex Paino, Nikolas Tezak, Jie Tang, Igor Babuschkin, Suchir Balaji, Shantanu Jain, William Saunders, Christopher Hesse, Andrew~N. Carr, Jan Leike, Josh Achiam, Vedant Misra, Evan Morikawa, Alec Radford, Matthew Knight, Miles Brundage, Mira Murati, Katie Mayer, Peter Welinder, Bob McGrew, Dario Amodei, Sam McCandlish, Ilya Sutskever, and Wojciech Zaremba.
\newblock Evaluating large language models trained on code, 2021.
\newblock URL \url{https://arxiv.org/abs/2107.03374}.

\bibitem[Cobbe et~al.(2021)Cobbe, Kosaraju, Bavarian, Chen, Jun, Kaiser, Plappert, Tworek, Hilton, Nakano, Hesse, and Schulman]{cobbe2021trainingverifierssolvemath}
Karl Cobbe, Vineet Kosaraju, Mohammad Bavarian, Mark Chen, Heewoo Jun, Lukasz Kaiser, Matthias Plappert, Jerry Tworek, Jacob Hilton, Reiichiro Nakano, Christopher Hesse, and John Schulman.
\newblock Training verifiers to solve math word problems, 2021.
\newblock URL \url{https://arxiv.org/abs/2110.14168}.

\bibitem[DeepSeek-AI et~al.(2024)DeepSeek-AI, Liu, Feng, Wang, Wang, Liu, Zhao, Dengr, Ruan, Dai, Guo, Yang, Chen, Ji, Li, Lin, Luo, Hao, Chen, Li, Zhang, Xu, Yang, Zhang, Ding, Xin, Gao, Li, Qu, Cai, Liang, Guo, Ni, Li, Chen, Yuan, Qiu, Song, Dong, Gao, Guan, Wang, Zhang, Xu, Xia, Zhao, Zhang, Li, Wang, Zhang, Zhang, Tang, Li, Tian, Huang, Wang, Zhang, Zhu, Chen, Du, Chen, Jin, Ge, Pan, Xu, Chen, Li, Lu, Zhou, Chen, Wu, Ye, Ma, Wang, Zhou, Yu, Zhou, Zheng, Wang, Pei, Yuan, Sun, Xiao, Zeng, An, Liu, Liang, Gao, Zhang, Li, Jin, Wang, Bi, Liu, Wang, Shen, Chen, Chen, Nie, Sun, Wang, Liu, Xie, Yu, Song, Zhou, Yang, Lu, Su, Wu, Li, Wei, Zhu, Xu, Huang, Li, Zhao, Sun, Li, Wang, Zheng, Zhang, Xiong, Zhao, He, Tang, Piao, Dong, Tan, Liu, Wang, Guo, Zhu, Wang, Zou, Zha, Ma, Yan, You, Liu, Ren, Ren, Sha, Fu, Huang, Zhang, Xie, Hao, Shao, Wen, Xu, Zhang, Li, Wang, Gu, Li, and Xie]{deepseekai2024deepseekv2strongeconomicalefficient}
DeepSeek-AI, Aixin Liu, Bei Feng, Bin Wang, Bingxuan Wang, Bo~Liu, Chenggang Zhao, Chengqi Dengr, Chong Ruan, Damai Dai, Daya Guo, Dejian Yang, Deli Chen, Dongjie Ji, Erhang Li, Fangyun Lin, Fuli Luo, Guangbo Hao, Guanting Chen, Guowei Li, H.~Zhang, Hanwei Xu, Hao Yang, Haowei Zhang, Honghui Ding, Huajian Xin, Huazuo Gao, Hui Li, Hui Qu, J.~L. Cai, Jian Liang, Jianzhong Guo, Jiaqi Ni, Jiashi Li, Jin Chen, Jingyang Yuan, Junjie Qiu, Junxiao Song, Kai Dong, Kaige Gao, Kang Guan, Lean Wang, Lecong Zhang, Lei Xu, Leyi Xia, Liang Zhao, Liyue Zhang, Meng Li, Miaojun Wang, Mingchuan Zhang, Minghua Zhang, Minghui Tang, Mingming Li, Ning Tian, Panpan Huang, Peiyi Wang, Peng Zhang, Qihao Zhu, Qinyu Chen, Qiushi Du, R.~J. Chen, R.~L. Jin, Ruiqi Ge, Ruizhe Pan, Runxin Xu, Ruyi Chen, S.~S. Li, Shanghao Lu, Shangyan Zhou, Shanhuang Chen, Shaoqing Wu, Shengfeng Ye, Shirong Ma, Shiyu Wang, Shuang Zhou, Shuiping Yu, Shunfeng Zhou, Size Zheng, T.~Wang, Tian Pei, Tian Yuan, Tianyu Sun, W.~L. Xiao, Wangding Zeng, Wei An, Wen
  Liu, Wenfeng Liang, Wenjun Gao, Wentao Zhang, X.~Q. Li, Xiangyue Jin, Xianzu Wang, Xiao Bi, Xiaodong Liu, Xiaohan Wang, Xiaojin Shen, Xiaokang Chen, Xiaosha Chen, Xiaotao Nie, Xiaowen Sun, Xiaoxiang Wang, Xin Liu, Xin Xie, Xingkai Yu, Xinnan Song, Xinyi Zhou, Xinyu Yang, Xuan Lu, Xuecheng Su, Y.~Wu, Y.~K. Li, Y.~X. Wei, Y.~X. Zhu, Yanhong Xu, Yanping Huang, Yao Li, Yao Zhao, Yaofeng Sun, Yaohui Li, Yaohui Wang, Yi~Zheng, Yichao Zhang, Yiliang Xiong, Yilong Zhao, Ying He, Ying Tang, Yishi Piao, Yixin Dong, Yixuan Tan, Yiyuan Liu, Yongji Wang, Yongqiang Guo, Yuchen Zhu, Yuduan Wang, Yuheng Zou, Yukun Zha, Yunxian Ma, Yuting Yan, Yuxiang You, Yuxuan Liu, Z.~Z. Ren, Zehui Ren, Zhangli Sha, Zhe Fu, Zhen Huang, Zhen Zhang, Zhenda Xie, Zhewen Hao, Zhihong Shao, Zhiniu Wen, Zhipeng Xu, Zhongyu Zhang, Zhuoshu Li, Zihan Wang, Zihui Gu, Zilin Li, and Ziwei Xie.
\newblock Deepseek-v2: A strong, economical, and efficient mixture-of-experts language model, 2024.
\newblock URL \url{https://arxiv.org/abs/2405.04434}.

\bibitem[Dong et~al.(2024)Dong, Jiang, Jin, and Li]{dong2024selfcollaborationcodegenerationchatgpt}
Yihong Dong, Xue Jiang, Zhi Jin, and Ge~Li.
\newblock Self-collaboration code generation via chatgpt, 2024.
\newblock URL \url{https://arxiv.org/abs/2304.07590}.

\bibitem[Dubey et~al.(2024)Dubey, Jauhri, Pandey, Kadian, Al-Dahle, Letman, Mathur, Schelten, Yang, Fan, et~al.]{dubey2024llama}
Abhimanyu Dubey, Abhinav Jauhri, Abhinav Pandey, Abhishek Kadian, Ahmad Al-Dahle, Aiesha Letman, Akhil Mathur, Alan Schelten, Amy Yang, Angela Fan, et~al.
\newblock The llama 3 herd of models.
\newblock \emph{arXiv preprint arXiv:2407.21783}, 2024.

\bibitem[GLM et~al.(2024)GLM, Zeng, Xu, Wang, Zhang, Yin, Rojas, Feng, Zhao, Lai, Yu, Wang, Sun, Zhang, Cheng, Gui, Tang, Zhang, Li, Zhao, Wu, Zhong, Liu, Huang, Zhang, Zheng, Lu, Duan, Zhang, Cao, Yang, Tam, Zhao, Liu, Xia, Zhang, Gu, Lv, Liu, Liu, Yang, Song, Zhang, An, Xu, Niu, Yang, Li, Bai, Dong, Qi, Wang, Yang, Du, Hou, and Wang]{glm2024chatglm}
Team GLM, Aohan Zeng, Bin Xu, Bowen Wang, Chenhui Zhang, Da~Yin, Diego Rojas, Guanyu Feng, Hanlin Zhao, Hanyu Lai, Hao Yu, Hongning Wang, Jiadai Sun, Jiajie Zhang, Jiale Cheng, Jiayi Gui, Jie Tang, Jing Zhang, Juanzi Li, Lei Zhao, Lindong Wu, Lucen Zhong, Mingdao Liu, Minlie Huang, Peng Zhang, Qinkai Zheng, Rui Lu, Shuaiqi Duan, Shudan Zhang, Shulin Cao, Shuxun Yang, Weng~Lam Tam, Wenyi Zhao, Xiao Liu, Xiao Xia, Xiaohan Zhang, Xiaotao Gu, Xin Lv, Xinghan Liu, Xinyi Liu, Xinyue Yang, Xixuan Song, Xunkai Zhang, Yifan An, Yifan Xu, Yilin Niu, Yuantao Yang, Yueyan Li, Yushi Bai, Yuxiao Dong, Zehan Qi, Zhaoyu Wang, Zhen Yang, Zhengxiao Du, Zhenyu Hou, and Zihan Wang.
\newblock Chatglm: A family of large language models from glm-130b to glm-4 all tools, 2024.

\bibitem[He-Yueya et~al.(2023)He-Yueya, Poesia, Wang, and Goodman]{heyueya2023solvingmathwordproblems}
Joy He-Yueya, Gabriel Poesia, Rose~E. Wang, and Noah~D. Goodman.
\newblock Solving math word problems by combining language models with symbolic solvers, 2023.
\newblock URL \url{https://arxiv.org/abs/2304.09102}.

\bibitem[Hendrycks et~al.(2021)Hendrycks, Burns, Kadavath, Arora, Basart, Tang, Song, and Steinhardt]{hendrycks2021measuringmathematicalproblemsolving}
Dan Hendrycks, Collin Burns, Saurav Kadavath, Akul Arora, Steven Basart, Eric Tang, Dawn Song, and Jacob Steinhardt.
\newblock Measuring mathematical problem solving with the math dataset, 2021.
\newblock URL \url{https://arxiv.org/abs/2103.03874}.

\bibitem[Imani et~al.(2023)Imani, Du, and Shrivastava]{imani2023mathpromptermathematicalreasoningusing}
Shima Imani, Liang Du, and Harsh Shrivastava.
\newblock Mathprompter: Mathematical reasoning using large language models, 2023.
\newblock URL \url{https://arxiv.org/abs/2303.05398}.

\bibitem[Lewkowycz et~al.(2022)Lewkowycz, Andreassen, Dohan, Dyer, Michalewski, Ramasesh, Slone, Anil, Schlag, Gutman-Solo, Wu, Neyshabur, Gur-Ari, and Misra]{lewkowycz2022solvingquantitativereasoningproblems}
Aitor Lewkowycz, Anders Andreassen, David Dohan, Ethan Dyer, Henryk Michalewski, Vinay Ramasesh, Ambrose Slone, Cem Anil, Imanol Schlag, Theo Gutman-Solo, Yuhuai Wu, Behnam Neyshabur, Guy Gur-Ari, and Vedant Misra.
\newblock Solving quantitative reasoning problems with language models, 2022.
\newblock URL \url{https://arxiv.org/abs/2206.14858}.

\bibitem[Li et~al.(2024)Li, Wang, Hu, Wei, Zheng, Hu, Zhang, and Peng]{li2024common7blanguagemodels}
Chen Li, Weiqi Wang, Jingcheng Hu, Yixuan Wei, Nanning Zheng, Han Hu, Zheng Zhang, and Houwen Peng.
\newblock Common 7b language models already possess strong math capabilities, 2024.
\newblock URL \url{https://arxiv.org/abs/2403.04706}.

\bibitem[Li et~al.(2023{\natexlab{a}})Li, Allal, Zi, Muennighoff, Kocetkov, Mou, Marone, Akiki, Li, Chim, Liu, Zheltonozhskii, Zhuo, Wang, Dehaene, Davaadorj, Lamy-Poirier, Monteiro, Shliazhko, Gontier, Meade, Zebaze, Yee, Umapathi, Zhu, Lipkin, Oblokulov, Wang, Murthy, Stillerman, Patel, Abulkhanov, Zocca, Dey, Zhang, Fahmy, Bhattacharyya, Yu, Singh, Luccioni, Villegas, Kunakov, Zhdanov, Romero, Lee, Timor, Ding, Schlesinger, Schoelkopf, Ebert, Dao, Mishra, Gu, Robinson, Anderson, Dolan-Gavitt, Contractor, Reddy, Fried, Bahdanau, Jernite, Ferrandis, Hughes, Wolf, Guha, von Werra, and de~Vries]{li2023starcodersourceyou}
Raymond Li, Loubna~Ben Allal, Yangtian Zi, Niklas Muennighoff, Denis Kocetkov, Chenghao Mou, Marc Marone, Christopher Akiki, Jia Li, Jenny Chim, Qian Liu, Evgenii Zheltonozhskii, Terry~Yue Zhuo, Thomas Wang, Olivier Dehaene, Mishig Davaadorj, Joel Lamy-Poirier, João Monteiro, Oleh Shliazhko, Nicolas Gontier, Nicholas Meade, Armel Zebaze, Ming-Ho Yee, Logesh~Kumar Umapathi, Jian Zhu, Benjamin Lipkin, Muhtasham Oblokulov, Zhiruo Wang, Rudra Murthy, Jason Stillerman, Siva~Sankalp Patel, Dmitry Abulkhanov, Marco Zocca, Manan Dey, Zhihan Zhang, Nour Fahmy, Urvashi Bhattacharyya, Wenhao Yu, Swayam Singh, Sasha Luccioni, Paulo Villegas, Maxim Kunakov, Fedor Zhdanov, Manuel Romero, Tony Lee, Nadav Timor, Jennifer Ding, Claire Schlesinger, Hailey Schoelkopf, Jan Ebert, Tri Dao, Mayank Mishra, Alex Gu, Jennifer Robinson, Carolyn~Jane Anderson, Brendan Dolan-Gavitt, Danish Contractor, Siva Reddy, Daniel Fried, Dzmitry Bahdanau, Yacine Jernite, Carlos~Muñoz Ferrandis, Sean Hughes, Thomas Wolf, Arjun Guha, Leandro von
  Werra, and Harm de~Vries.
\newblock Starcoder: may the source be with you!, 2023{\natexlab{a}}.
\newblock URL \url{https://arxiv.org/abs/2305.06161}.

\bibitem[Li et~al.(2023{\natexlab{b}})Li, Lin, Zhang, Fu, Chen, Lou, and Chen]{li-etal-2023-making}
Yifei Li, Zeqi Lin, Shizhuo Zhang, Qiang Fu, Bei Chen, Jian-Guang Lou, and Weizhu Chen.
\newblock Making language models better reasoners with step-aware verifier.
\newblock In Anna Rogers, Jordan Boyd-Graber, and Naoaki Okazaki (eds.), \emph{Proceedings of the 61st Annual Meeting of the Association for Computational Linguistics (Volume 1: Long Papers)}, pp.\  5315--5333, Toronto, Canada, July 2023{\natexlab{b}}. Association for Computational Linguistics.
\newblock \doi{10.18653/v1/2023.acl-long.291}.
\newblock URL \url{https://aclanthology.org/2023.acl-long.291}.

\bibitem[Ling et~al.(2017)Ling, Yogatama, Dyer, and Blunsom]{ling-etal-2017-program}
Wang Ling, Dani Yogatama, Chris Dyer, and Phil Blunsom.
\newblock Program induction by rationale generation: Learning to solve and explain algebraic word problems.
\newblock In Regina Barzilay and Min-Yen Kan (eds.), \emph{Proceedings of the 55th Annual Meeting of the Association for Computational Linguistics (Volume 1: Long Papers)}, pp.\  158--167, Vancouver, Canada, July 2017. Association for Computational Linguistics.
\newblock \doi{10.18653/v1/P17-1015}.
\newblock URL \url{https://aclanthology.org/P17-1015}.

\bibitem[Luo et~al.(2023)Luo, Sun, Xu, Zhao, Lou, Tao, Geng, Lin, Chen, and Zhang]{luo2023wizardmathempoweringmathematicalreasoning}
Haipeng Luo, Qingfeng Sun, Can Xu, Pu~Zhao, Jianguang Lou, Chongyang Tao, Xiubo Geng, Qingwei Lin, Shifeng Chen, and Dongmei Zhang.
\newblock Wizardmath: Empowering mathematical reasoning for large language models via reinforced evol-instruct, 2023.
\newblock URL \url{https://arxiv.org/abs/2308.09583}.

\bibitem[Mishra et~al.(2022)Mishra, Finlayson, Lu, Tang, Welleck, Baral, Rajpurohit, Tafjord, Sabharwal, Clark, and Kalyan]{mishra-etal-2022-lila}
Swaroop Mishra, Matthew Finlayson, Pan Lu, Leonard Tang, Sean Welleck, Chitta Baral, Tanmay Rajpurohit, Oyvind Tafjord, Ashish Sabharwal, Peter Clark, and Ashwin Kalyan.
\newblock {LILA}: A unified benchmark for mathematical reasoning.
\newblock In Yoav Goldberg, Zornitsa Kozareva, and Yue Zhang (eds.), \emph{Proceedings of the 2022 Conference on Empirical Methods in Natural Language Processing}, pp.\  5807--5832, Abu Dhabi, United Arab Emirates, December 2022. Association for Computational Linguistics.
\newblock \doi{10.18653/v1/2022.emnlp-main.392}.
\newblock URL \url{https://aclanthology.org/2022.emnlp-main.392}.

\bibitem[Nye et~al.(2021)Nye, Andreassen, Gur-Ari, Michalewski, Austin, Bieber, Dohan, Lewkowycz, Bosma, Luan, Sutton, and Odena]{nye2021workscratchpadsintermediatecomputation}
Maxwell Nye, Anders~Johan Andreassen, Guy Gur-Ari, Henryk Michalewski, Jacob Austin, David Bieber, David Dohan, Aitor Lewkowycz, Maarten Bosma, David Luan, Charles Sutton, and Augustus Odena.
\newblock Show your work: Scratchpads for intermediate computation with language models, 2021.
\newblock URL \url{https://arxiv.org/abs/2112.00114}.

\bibitem[OpenAI(2024{\natexlab{a}})]{gpt4o}
OpenAI, 2024{\natexlab{a}}.
\newblock URL \url{https://openai.com/index/hello-gpt-4o}.

\bibitem[OpenAI(2024{\natexlab{b}})]{gpt4omini}
OpenAI, 2024{\natexlab{b}}.
\newblock URL \url{https://openai.com/index/gpt-4o-mini-advancing-cost-efficient-intelligence}.

\bibitem[OpenAI(2024{\natexlab{c}})]{o1}
OpenAI, 2024{\natexlab{c}}.
\newblock URL \url{https://openai.com/index/learning-to-reason-with-llms/}.

\bibitem[OpenAI et~al.(2024)OpenAI, Achiam, Adler, Agarwal, Ahmad, Akkaya, Aleman, Almeida, Altenschmidt, Altman, Anadkat, Avila, Babuschkin, Balaji, Balcom, Baltescu, Bao, Bavarian, Belgum, Bello, Berdine, Bernadett-Shapiro, Berner, Bogdonoff, Boiko, Boyd, Brakman, Brockman, Brooks, Brundage, Button, Cai, Campbell, Cann, Carey, Carlson, Carmichael, Chan, Chang, Chantzis, Chen, Chen, Chen, Chen, Chen, Chess, Cho, Chu, Chung, Cummings, Currier, Dai, Decareaux, Degry, Deutsch, Deville, Dhar, Dohan, Dowling, Dunning, Ecoffet, Eleti, Eloundou, Farhi, Fedus, Felix, Fishman, Forte, Fulford, Gao, Georges, Gibson, Goel, Gogineni, Goh, Gontijo-Lopes, Gordon, Grafstein, Gray, Greene, Gross, Gu, Guo, Hallacy, Han, Harris, He, Heaton, Heidecke, Hesse, Hickey, Hickey, Hoeschele, Houghton, Hsu, Hu, Hu, Huizinga, Jain, Jain, Jang, Jiang, Jiang, Jin, Jin, Jomoto, Jonn, Jun, Kaftan, Łukasz Kaiser, Kamali, Kanitscheider, Keskar, Khan, Kilpatrick, Kim, Kim, Kim, Kirchner, Kiros, Knight, Kokotajlo, Łukasz Kondraciuk, Kondrich,
  Konstantinidis, Kosic, Krueger, Kuo, Lampe, Lan, Lee, Leike, Leung, Levy, Li, Lim, Lin, Lin, Litwin, Lopez, Lowe, Lue, Makanju, Malfacini, Manning, Markov, Markovski, Martin, Mayer, Mayne, McGrew, McKinney, McLeavey, McMillan, McNeil, Medina, Mehta, Menick, Metz, Mishchenko, Mishkin, Monaco, Morikawa, Mossing, Mu, Murati, Murk, Mély, Nair, Nakano, Nayak, Neelakantan, Ngo, Noh, Ouyang, O'Keefe, Pachocki, Paino, Palermo, Pantuliano, Parascandolo, Parish, Parparita, Passos, Pavlov, Peng, Perelman, de~Avila Belbute~Peres, Petrov, de~Oliveira~Pinto, Michael, Pokorny, Pokrass, Pong, Powell, Power, Power, Proehl, Puri, Radford, Rae, Ramesh, Raymond, Real, Rimbach, Ross, Rotsted, Roussez, Ryder, Saltarelli, Sanders, Santurkar, Sastry, Schmidt, Schnurr, Schulman, Selsam, Sheppard, Sherbakov, Shieh, Shoker, Shyam, Sidor, Sigler, Simens, Sitkin, Slama, Sohl, Sokolowsky, Song, Staudacher, Such, Summers, Sutskever, Tang, Tezak, Thompson, Tillet, Tootoonchian, Tseng, Tuggle, Turley, Tworek, Uribe, Vallone, Vijayvergiya,
  Voss, Wainwright, Wang, Wang, Wang, Ward, Wei, Weinmann, Welihinda, Welinder, Weng, Weng, Wiethoff, Willner, Winter, Wolrich, Wong, Workman, Wu, Wu, Wu, Xiao, Xu, Yoo, Yu, Yuan, Zaremba, Zellers, Zhang, Zhang, Zhao, Zheng, Zhuang, Zhuk, and Zoph]{openai2024gpt4technicalreport}
OpenAI, Josh Achiam, Steven Adler, Sandhini Agarwal, Lama Ahmad, Ilge Akkaya, Florencia~Leoni Aleman, Diogo Almeida, Janko Altenschmidt, Sam Altman, Shyamal Anadkat, Red Avila, Igor Babuschkin, Suchir Balaji, Valerie Balcom, Paul Baltescu, Haiming Bao, Mohammad Bavarian, Jeff Belgum, Irwan Bello, Jake Berdine, Gabriel Bernadett-Shapiro, Christopher Berner, Lenny Bogdonoff, Oleg Boiko, Madelaine Boyd, Anna-Luisa Brakman, Greg Brockman, Tim Brooks, Miles Brundage, Kevin Button, Trevor Cai, Rosie Campbell, Andrew Cann, Brittany Carey, Chelsea Carlson, Rory Carmichael, Brooke Chan, Che Chang, Fotis Chantzis, Derek Chen, Sully Chen, Ruby Chen, Jason Chen, Mark Chen, Ben Chess, Chester Cho, Casey Chu, Hyung~Won Chung, Dave Cummings, Jeremiah Currier, Yunxing Dai, Cory Decareaux, Thomas Degry, Noah Deutsch, Damien Deville, Arka Dhar, David Dohan, Steve Dowling, Sheila Dunning, Adrien Ecoffet, Atty Eleti, Tyna Eloundou, David Farhi, Liam Fedus, Niko Felix, Simón~Posada Fishman, Juston Forte, Isabella Fulford, Leo
  Gao, Elie Georges, Christian Gibson, Vik Goel, Tarun Gogineni, Gabriel Goh, Rapha Gontijo-Lopes, Jonathan Gordon, Morgan Grafstein, Scott Gray, Ryan Greene, Joshua Gross, Shixiang~Shane Gu, Yufei Guo, Chris Hallacy, Jesse Han, Jeff Harris, Yuchen He, Mike Heaton, Johannes Heidecke, Chris Hesse, Alan Hickey, Wade Hickey, Peter Hoeschele, Brandon Houghton, Kenny Hsu, Shengli Hu, Xin Hu, Joost Huizinga, Shantanu Jain, Shawn Jain, Joanne Jang, Angela Jiang, Roger Jiang, Haozhun Jin, Denny Jin, Shino Jomoto, Billie Jonn, Heewoo Jun, Tomer Kaftan, Łukasz Kaiser, Ali Kamali, Ingmar Kanitscheider, Nitish~Shirish Keskar, Tabarak Khan, Logan Kilpatrick, Jong~Wook Kim, Christina Kim, Yongjik Kim, Jan~Hendrik Kirchner, Jamie Kiros, Matt Knight, Daniel Kokotajlo, Łukasz Kondraciuk, Andrew Kondrich, Aris Konstantinidis, Kyle Kosic, Gretchen Krueger, Vishal Kuo, Michael Lampe, Ikai Lan, Teddy Lee, Jan Leike, Jade Leung, Daniel Levy, Chak~Ming Li, Rachel Lim, Molly Lin, Stephanie Lin, Mateusz Litwin, Theresa Lopez, Ryan
  Lowe, Patricia Lue, Anna Makanju, Kim Malfacini, Sam Manning, Todor Markov, Yaniv Markovski, Bianca Martin, Katie Mayer, Andrew Mayne, Bob McGrew, Scott~Mayer McKinney, Christine McLeavey, Paul McMillan, Jake McNeil, David Medina, Aalok Mehta, Jacob Menick, Luke Metz, Andrey Mishchenko, Pamela Mishkin, Vinnie Monaco, Evan Morikawa, Daniel Mossing, Tong Mu, Mira Murati, Oleg Murk, David Mély, Ashvin Nair, Reiichiro Nakano, Rajeev Nayak, Arvind Neelakantan, Richard Ngo, Hyeonwoo Noh, Long Ouyang, Cullen O'Keefe, Jakub Pachocki, Alex Paino, Joe Palermo, Ashley Pantuliano, Giambattista Parascandolo, Joel Parish, Emy Parparita, Alex Passos, Mikhail Pavlov, Andrew Peng, Adam Perelman, Filipe de~Avila Belbute~Peres, Michael Petrov, Henrique~Ponde de~Oliveira~Pinto, Michael, Pokorny, Michelle Pokrass, Vitchyr~H. Pong, Tolly Powell, Alethea Power, Boris Power, Elizabeth Proehl, Raul Puri, Alec Radford, Jack Rae, Aditya Ramesh, Cameron Raymond, Francis Real, Kendra Rimbach, Carl Ross, Bob Rotsted, Henri Roussez,
  Nick Ryder, Mario Saltarelli, Ted Sanders, Shibani Santurkar, Girish Sastry, Heather Schmidt, David Schnurr, John Schulman, Daniel Selsam, Kyla Sheppard, Toki Sherbakov, Jessica Shieh, Sarah Shoker, Pranav Shyam, Szymon Sidor, Eric Sigler, Maddie Simens, Jordan Sitkin, Katarina Slama, Ian Sohl, Benjamin Sokolowsky, Yang Song, Natalie Staudacher, Felipe~Petroski Such, Natalie Summers, Ilya Sutskever, Jie Tang, Nikolas Tezak, Madeleine~B. Thompson, Phil Tillet, Amin Tootoonchian, Elizabeth Tseng, Preston Tuggle, Nick Turley, Jerry Tworek, Juan Felipe~Cerón Uribe, Andrea Vallone, Arun Vijayvergiya, Chelsea Voss, Carroll Wainwright, Justin~Jay Wang, Alvin Wang, Ben Wang, Jonathan Ward, Jason Wei, CJ~Weinmann, Akila Welihinda, Peter Welinder, Jiayi Weng, Lilian Weng, Matt Wiethoff, Dave Willner, Clemens Winter, Samuel Wolrich, Hannah Wong, Lauren Workman, Sherwin Wu, Jeff Wu, Michael Wu, Kai Xiao, Tao Xu, Sarah Yoo, Kevin Yu, Qiming Yuan, Wojciech Zaremba, Rowan Zellers, Chong Zhang, Marvin Zhang, Shengjia
  Zhao, Tianhao Zheng, Juntang Zhuang, William Zhuk, and Barret Zoph.
\newblock Gpt-4 technical report, 2024.
\newblock URL \url{https://arxiv.org/abs/2303.08774}.

\bibitem[Ouyang et~al.(2022)Ouyang, Wu, Jiang, Almeida, Wainwright, Mishkin, Zhang, Agarwal, Slama, Ray, Schulman, Hilton, Kelton, Miller, Simens, Askell, Welinder, Christiano, Leike, and Lowe]{ouyang2022traininglanguagemodelsfollow}
Long Ouyang, Jeff Wu, Xu~Jiang, Diogo Almeida, Carroll~L. Wainwright, Pamela Mishkin, Chong Zhang, Sandhini Agarwal, Katarina Slama, Alex Ray, John Schulman, Jacob Hilton, Fraser Kelton, Luke Miller, Maddie Simens, Amanda Askell, Peter Welinder, Paul Christiano, Jan Leike, and Ryan Lowe.
\newblock Training language models to follow instructions with human feedback, 2022.
\newblock URL \url{https://arxiv.org/abs/2203.02155}.

\bibitem[Qian et~al.(2024)Qian, Liu, Liu, Chen, Dang, Li, Yang, Chen, Su, Cong, Xu, Li, Liu, and Sun]{qian2024chatdevcommunicativeagentssoftware}
Chen Qian, Wei Liu, Hongzhang Liu, Nuo Chen, Yufan Dang, Jiahao Li, Cheng Yang, Weize Chen, Yusheng Su, Xin Cong, Juyuan Xu, Dahai Li, Zhiyuan Liu, and Maosong Sun.
\newblock Chatdev: Communicative agents for software development, 2024.
\newblock URL \url{https://arxiv.org/abs/2307.07924}.

\bibitem[Raiyan et~al.(2023)Raiyan, Faiyaz, Kabir, Kabir, Mahmud, and Hasan]{raiyan-etal-2023-math}
Syed~Rifat Raiyan, Md~Nafis Faiyaz, Shah Md.~Jawad Kabir, Mohsinul Kabir, Hasan Mahmud, and Md~Kamrul Hasan.
\newblock Math word problem solving by generating linguistic variants of problem statements.
\newblock In Vishakh Padmakumar, Gisela Vallejo, and Yao Fu (eds.), \emph{Proceedings of the 61st Annual Meeting of the Association for Computational Linguistics (Volume 4: Student Research Workshop)}, pp.\  362--378, Toronto, Canada, July 2023. Association for Computational Linguistics.
\newblock \doi{10.18653/v1/2023.acl-srw.49}.
\newblock URL \url{https://aclanthology.org/2023.acl-srw.49}.

\bibitem[Reid et~al.(2024)Reid, Savinov, Teplyashin, Lepikhin, Lillicrap, Alayrac, Soricut, Lazaridou, Firat, Schrittwieser, et~al.]{reid2024gemini}
Machel Reid, Nikolay Savinov, Denis Teplyashin, Dmitry Lepikhin, Timothy Lillicrap, Jean-baptiste Alayrac, Radu Soricut, Angeliki Lazaridou, Orhan Firat, Julian Schrittwieser, et~al.
\newblock Gemini 1.5: Unlocking multimodal understanding across millions of tokens of context.
\newblock \emph{arXiv preprint arXiv:2403.05530}, 2024.

\bibitem[Rozière et~al.(2024)Rozière, Gehring, Gloeckle, Sootla, Gat, Tan, Adi, Liu, Sauvestre, Remez, Rapin, Kozhevnikov, Evtimov, Bitton, Bhatt, Ferrer, Grattafiori, Xiong, Défossez, Copet, Azhar, Touvron, Martin, Usunier, Scialom, and Synnaeve]{rozière2024codellamaopenfoundation}
Baptiste Rozière, Jonas Gehring, Fabian Gloeckle, Sten Sootla, Itai Gat, Xiaoqing~Ellen Tan, Yossi Adi, Jingyu Liu, Romain Sauvestre, Tal Remez, Jérémy Rapin, Artyom Kozhevnikov, Ivan Evtimov, Joanna Bitton, Manish Bhatt, Cristian~Canton Ferrer, Aaron Grattafiori, Wenhan Xiong, Alexandre Défossez, Jade Copet, Faisal Azhar, Hugo Touvron, Louis Martin, Nicolas Usunier, Thomas Scialom, and Gabriel Synnaeve.
\newblock Code llama: Open foundation models for code, 2024.
\newblock URL \url{https://arxiv.org/abs/2308.12950}.

\bibitem[Saxton et~al.(2019)Saxton, Grefenstette, Hill, and Kohli]{saxton2019analysing}
David Saxton, Edward Grefenstette, Felix Hill, and Pushmeet Kohli.
\newblock Analysing mathematical reasoning abilities of neural models.
\newblock \emph{arXiv preprint arXiv:1904.01557}, 2019.

\bibitem[Shinn et~al.(2023)Shinn, Cassano, Berman, Gopinath, Narasimhan, and Yao]{shinn2023reflexionlanguageagentsverbal}
Noah Shinn, Federico Cassano, Edward Berman, Ashwin Gopinath, Karthik Narasimhan, and Shunyu Yao.
\newblock Reflexion: Language agents with verbal reinforcement learning, 2023.
\newblock URL \url{https://arxiv.org/abs/2303.11366}.

\bibitem[Tang et~al.(2024{\natexlab{a}})Tang, Zou, Zhang, Li, Zhao, Zhang, Cohan, and Gerstein]{tang-etal-2024-medagents}
Xiangru Tang, Anni Zou, Zhuosheng Zhang, Ziming Li, Yilun Zhao, Xingyao Zhang, Arman Cohan, and Mark Gerstein.
\newblock {M}ed{A}gents: Large language models as collaborators for zero-shot medical reasoning.
\newblock In Lun-Wei Ku, Andre Martins, and Vivek Srikumar (eds.), \emph{Findings of the Association for Computational Linguistics ACL 2024}, pp.\  599--621, Bangkok, Thailand and virtual meeting, August 2024{\natexlab{a}}. Association for Computational Linguistics.
\newblock URL \url{https://aclanthology.org/2024.findings-acl.33}.

\bibitem[Tang et~al.(2024{\natexlab{b}})Tang, Zhang, Wang, and Wei]{tang2024mathscalescalinginstructiontuning}
Zhengyang Tang, Xingxing Zhang, Benyou Wang, and Furu Wei.
\newblock Mathscale: Scaling instruction tuning for mathematical reasoning, 2024{\natexlab{b}}.
\newblock URL \url{https://arxiv.org/abs/2403.02884}.

\bibitem[Taori et~al.(2023)Taori, Gulrajani, Zhang, Dubois, Li, Guestrin, Liang, and Hashimoto]{alpaca}
Rohan Taori, Ishaan Gulrajani, Tianyi Zhang, Yann Dubois, Xuechen Li, Carlos Guestrin, Percy Liang, and Tatsunori~B. Hashimoto.
\newblock Stanford alpaca: An instruction-following llama model.
\newblock \url{https://github.com/tatsu-lab/stanford_alpaca}, 2023.

\bibitem[Team(2024)]{qwen2.5}
Qwen Team.
\newblock Qwen2.5: A party of foundation models, September 2024.
\newblock URL \url{https://qwenlm.github.io/blog/qwen2.5/}.

\bibitem[Tong et~al.(2024)Tong, Zhang, Wang, Wu, and He]{tong2024dartmathdifficultyawarerejectiontuning}
Yuxuan Tong, Xiwen Zhang, Rui Wang, Ruidong Wu, and Junxian He.
\newblock Dart-math: Difficulty-aware rejection tuning for mathematical problem-solving, 2024.
\newblock URL \url{https://arxiv.org/abs/2407.13690}.

\bibitem[Touvron et~al.(2023)Touvron, Martin, Stone, Albert, Almahairi, Babaei, Bashlykov, Batra, Bhargava, Bhosale, et~al.]{touvron2023llama}
Hugo Touvron, Louis Martin, Kevin Stone, Peter Albert, Amjad Almahairi, Yasmine Babaei, Nikolay Bashlykov, Soumya Batra, Prajjwal Bhargava, Shruti Bhosale, et~al.
\newblock Llama 2: Open foundation and fine-tuned chat models.
\newblock \emph{arXiv preprint arXiv:2307.09288}, 2023.

\bibitem[Wang et~al.(2023)Wang, Xie, Jiang, Mandlekar, Xiao, Zhu, Fan, and Anandkumar]{wang2023voyageropenendedembodiedagent}
Guanzhi Wang, Yuqi Xie, Yunfan Jiang, Ajay Mandlekar, Chaowei Xiao, Yuke Zhu, Linxi Fan, and Anima Anandkumar.
\newblock Voyager: An open-ended embodied agent with large language models, 2023.
\newblock URL \url{https://arxiv.org/abs/2305.16291}.

\bibitem[Wei et~al.(2023)Wei, Wang, Schuurmans, Bosma, Ichter, Xia, Chi, Le, and Zhou]{wei2023chainofthoughtpromptingelicitsreasoning}
Jason Wei, Xuezhi Wang, Dale Schuurmans, Maarten Bosma, Brian Ichter, Fei Xia, Ed~Chi, Quoc Le, and Denny Zhou.
\newblock Chain-of-thought prompting elicits reasoning in large language models, 2023.
\newblock URL \url{https://arxiv.org/abs/2201.11903}.

\bibitem[Wu et~al.(2024)Wu, Jia, Zhang, Li, Zhu, Wang, Lee, Peng, Wu, and Wang]{wu2024mathchatconversetacklechallenging}
Yiran Wu, Feiran Jia, Shaokun Zhang, Hangyu Li, Erkang Zhu, Yue Wang, Yin~Tat Lee, Richard Peng, Qingyun Wu, and Chi Wang.
\newblock Mathchat: Converse to tackle challenging math problems with llm agents, 2024.
\newblock URL \url{https://arxiv.org/abs/2306.01337}.

\bibitem[Yamauchi et~al.(2023)Yamauchi, Sonoda, Sannai, and Kumagai]{yamauchi2023lpmlllmpromptingmarkuplanguage}
Ryutaro Yamauchi, Sho Sonoda, Akiyoshi Sannai, and Wataru Kumagai.
\newblock Lpml: Llm-prompting markup language for mathematical reasoning, 2023.
\newblock URL \url{https://arxiv.org/abs/2309.13078}.

\bibitem[Yang et~al.(2023)Yang, Ding, Lv, Jiang, He, Guo, Bai, and Tang]{yang2023gptsolvemathematicalproblems}
Zhen Yang, Ming Ding, Qingsong Lv, Zhihuan Jiang, Zehai He, Yuyi Guo, Jinfeng Bai, and Jie Tang.
\newblock Gpt can solve mathematical problems without a calculator, 2023.
\newblock URL \url{https://arxiv.org/abs/2309.03241}.

\bibitem[Yao et~al.(2023)Yao, Zhou, and Wang]{yao2023solvingmathwordproblem}
Jie Yao, Zihao Zhou, and Qiufeng Wang.
\newblock Solving math word problem with problem type classification, 2023.
\newblock URL \url{https://arxiv.org/abs/2308.13844}.

\bibitem[Yu et~al.(2024)Yu, Jiang, Shi, Yu, Liu, Zhang, Kwok, Li, Weller, and Liu]{yu2024metamathbootstrapmathematicalquestions}
Longhui Yu, Weisen Jiang, Han Shi, Jincheng Yu, Zhengying Liu, Yu~Zhang, James~T. Kwok, Zhenguo Li, Adrian Weller, and Weiyang Liu.
\newblock Metamath: Bootstrap your own mathematical questions for large language models, 2024.
\newblock URL \url{https://arxiv.org/abs/2309.12284}.

\bibitem[Yuan et~al.(2023)Yuan, Yuan, Li, Dong, Lu, Tan, Zhou, and Zhou]{yuan2023scalingrelationshiplearningmathematical}
Zheng Yuan, Hongyi Yuan, Chengpeng Li, Guanting Dong, Keming Lu, Chuanqi Tan, Chang Zhou, and Jingren Zhou.
\newblock Scaling relationship on learning mathematical reasoning with large language models, 2023.
\newblock URL \url{https://arxiv.org/abs/2308.01825}.

\bibitem[Yue et~al.(2023)Yue, Qu, Zhang, Fu, Huang, Sun, Su, and Chen]{yue2023mammothbuildingmathgeneralist}
Xiang Yue, Xingwei Qu, Ge~Zhang, Yao Fu, Wenhao Huang, Huan Sun, Yu~Su, and Wenhu Chen.
\newblock Mammoth: Building math generalist models through hybrid instruction tuning, 2023.
\newblock URL \url{https://arxiv.org/abs/2309.05653}.

\bibitem[Zhang et~al.(2024)Zhang, Hu, Zhoubian, Du, Yang, Wang, Yue, Dong, and Tang]{zhang2024sciglmtrainingscientificlanguage}
Dan Zhang, Ziniu Hu, Sining Zhoubian, Zhengxiao Du, Kaiyu Yang, Zihan Wang, Yisong Yue, Yuxiao Dong, and Jie Tang.
\newblock Sciglm: Training scientific language models with self-reflective instruction annotation and tuning, 2024.
\newblock URL \url{https://arxiv.org/abs/2401.07950}.

\bibitem[Zhang et~al.(2023)Zhang, Wang, Yang, Feng, and Lan]{zhang2023interpretablemathwordproblem}
Mengxue Zhang, Zichao Wang, Zhichao Yang, Weiqi Feng, and Andrew Lan.
\newblock Interpretable math word problem solution generation via step-by-step planning, 2023.
\newblock URL \url{https://arxiv.org/abs/2306.00784}.

\end{thebibliography}

\clearpage
\appendix
\begin{center}
    \bf\Large Appendix
\end{center}
\section{Experimental Details}\label{sec: exp}
\paragraph{Training Details.} For the entire training process, we use the AdamW optimizer with 3 epochs, leveraging 8 NVIDIA A800 80GB GPUs. The learning rate is set to 2e-5 with a 3\% learning rate warmup. We employed DeepSpeed for training, applying the ZeRO-2 stage. We set gradient accumulation steps to 2 with a per-device training batch size of 2. The max model lengths are set to 4096 for LLaMA-2-7B, LLaMA-3-8B and LLaMA-2-13B, while for Qwen2.5-7B, the max model length is also set to 2048 due to its maximum supported length. For more details, please refer to our released code.

\paragraph{Answer Extraction.} It's worth noting that when we prompt the closed-source LLM to generate the CoT solutions, we always include ``\textit{present the final answer enclosed in \textbackslash \textbackslash boxed\{\}}" at the end of the prompt. As a result, the answer will always be enclosed in a ``\textbackslash \textbackslash boxed\{\}", allowing us to simply extract the content within the ``\textbackslash \textbackslash boxed\{\}" as the answer. To further enhance the accuracy of answer extraction and comparison, we adopted the method used in DART-Math. For implementation details, please refer to their open-source code.

\section{Prompts}

\subsection{Prompt for Inference}\label{sub: Inference}
\begin{tcolorbox}[sharpish corners, boxrule = 0pt, toprule = 4.5pt, enhanced, fuzzy shadow = {0pt}{-2pt}{-0.5pt}{0.5pt}{black!35}, title=Prompt for Inference]
    \textit{Please provide a detailed, step-by-step explanation for the following math problem. At the end of the explanation, present the final answer enclosed in \textbackslash\textbackslash boxed\{\} \textbackslash n Math problem: }
\end{tcolorbox}

\subsection{Prompts for Classification}\label{sub: Classification}
\begin{tcolorbox}[sharpish corners, boxrule = 0pt, toprule = 4.5pt, enhanced, fuzzy shadow = {0pt}{-2pt}{-0.5pt}{0.5pt}{black!35}, title=Prompts for Classification]
    \textit{You are an expert in job classification according to the International Standard Classification of Occupations (ISCO-08). Given a description of a persona, classify their occupation into the closest ISCO-08 major group (e.g., \textbackslash"1 - Managers\textbackslash"). If the occupation cannot be identified, classify it as \textbackslash"Others.\textbackslash" \textbackslash n\textbackslash nPersona: \textbackslash n\{persona\}}
\end{tcolorbox}

\subsection{Prompt for Rewriting}\label{sub: Rewriting}
\begin{tcolorbox}[sharpish corners, boxrule = 0pt, toprule = 4.5pt, enhanced, fuzzy shadow = {0pt}{-2pt}{-0.5pt}{0.5pt}{black!35}, title=Prompt for Rewriting]
    \textit{Math problem: \{problem\} \textbackslash nPlease rephrase the above math problem with the following persona:\textbackslash n\{persona\}}
\end{tcolorbox}

\subsection{Prompt for Reflection}\label{sub: Reflection}
\begin{tcolorbox}[sharpish corners, boxrule = 0pt, toprule = 4.5pt, enhanced, fuzzy shadow = {0pt}{-2pt}{-0.5pt}{0.5pt}{black!35}, title=Prompt for Reflection]
    \textit{The following input consists of a math problem and a corresponding explanation. However, this explanation is incorrect, please reflect on its errors and then generate a corrected, detailed, step-by-step explanation for the following math problem. Divide your response into two parts: Review of Incorrect Explanation and Corrected Explanation. At the end of the explanation, present the final answer enclosed in \textbackslash \textbackslash boxed\{\}.\textbackslash nMath Problem: \{problem\}\textbackslash nIncorrect Explanation: \{explanation\}}
\end{tcolorbox}

\subsection{Prompt for Training}\label{sub: Training}
\begin{tcolorbox}[sharpish corners, boxrule = 0pt, toprule = 4.5pt, enhanced, fuzzy shadow = {0pt}{-2pt}{-0.5pt}{0.5pt}{black!35}, title=Training Prompt]
    \textit{Below is an instruction that describes a task. Write a response that appropriately completes the request.\textbackslash n\textbackslash n\#\#\# Instruction:\textbackslash n\{instruction\}\textbackslash n\textbackslash n\#\#\# Response:}
\end{tcolorbox}

\subsection{Prompt for Evaluation}\label{sub: Evaluation}
\begin{tcolorbox}[sharpish corners, boxrule = 0pt, toprule = 4.5pt, enhanced, fuzzy shadow = {0pt}{-2pt}{-0.5pt}{0.5pt}{black!35}, title=Evaluation Prompt]
    \textit{Below is an instruction that describes a task. Write a response that appropriately completes the request.\textbackslash n\textbackslash n\#\#\# Instruction:\textbackslash n\{instruction\}\textbackslash n\textbackslash n\#\#\# Response: Let’s think step by step.}
\end{tcolorbox}

\section{Detailed Composition of PersonaMathQA}\label{sec: compos}

\begin{table*}[!htbp]
\begin{center}
\scalebox{1}{
\begin{tabular}{c|cc|cc|c}
\toprule[1pt]
\multirow{2}{*}{Dataset} & \multicolumn{2}{c|}{Stage 1} &  \multicolumn{2}{c|}{Stage 2} & \multirow{2}{*}{Overall}\\
 & \multicolumn{1}{c|}{Inference} & Rewrite & \multicolumn{1}{c|}{Reflection} & Rewrite & \\
\hline
PersonaMathQA-GSM8K & \multicolumn{1}{c|}{6.6K} & 66.6K & \multicolumn{1}{c|}{0.1K} & 1.2K & 74.7K\\
\hline
PersonaMathQA-MATH & \multicolumn{1}{c|}{5.4K} & 46.4K & \multicolumn{1}{c|}{0.2K} & 2.0K & 54.2K\\
\hline
PersonaMathQA & \multicolumn{1}{c|}{12.1K} & 113.1K & \multicolumn{1}{c|}{0.3K} & 3.2K & 128.9K\\
\bottomrule
\end{tabular}}
\end{center}
\caption{The detailed composition of each component of the PersonaMathQA dataset. This table shows the number of questions with correct answers that were added to PersonaMathQA during the final step of each phase.}
\label{dataset}
\end{table*}

\clearpage

\section{Additional Case Studies}\label{sec: case}

\begin{tcolorbox}[sharpish corners, boxrule = 0pt, toprule = 4.5pt, enhanced, fuzzy shadow = {0pt}{-2pt}{-0.5pt}{0.5pt}{black!35}, title=Examples of complete responses from our LLM reflection framework]
\textbf{Example from MATH:}\\

\textit{Problem:} If $a$ and $b$ are positive integers for which $ab - 3a + 4b = 137$, what is the minimal possible value of $|a - b|$?\\

\textit{Incorrect Response:} We begin with the equation given in the problem: $ab - 3a + 4b = 137$. To analyze this equation, we can rearrange it to facilitate factoring: $ab - 3a + 4b - 137 = 0$. Next, we can rewrite the equation in a more manageable form, isolating \(ab\):$ab = 3a - 4b + 137$. Rearranging presents: $ab - 4b = 3a + 137$, $b(a - 4) = 3a + 137$. From this, we can solve for \(b\): $b = \frac{3a + 137}{a - 4}$. To ensure \(b\) is an integer, the term \(3a + 137\) must be divisible by \(a - 4\). We set up the condition for divisibility. To do this, we first simplify with polynomial long division (or by applying the remainder theorem). We can divide \(3a + 137\) by \(a - 4\): 1. The leading term calculates as \(3\). 2. Multiply \(3(a - 4)\):   $   3a - 12   $. 3. Subtract from the original expression:   $   (3a + 137) - (3a - 12) = 137 + 12 = 149   $. Thus, we can express this division as: $3a + 137 = 3(a - 4) + 149$. This indicates that the remainder, when dividing by \(a - 4\), is \(149\). Therefore, for \(b\) to be an integer, \(149\) must be divisible by \(a - 4\). This leads us to find the positive divisors of \(149\). The divisors of \(149\) (which is a prime number) are: $1 \quad \text{and} \quad 149$. Now we analyze each case:\#\#\# Case 1: \(a - 4 = 1\)$a = 5$. Substituting into our equation for \(b\):$b = \frac{3(5) + 137}{5 - 4} = \frac{15 + 137}{1} = 152$. Here, \(a = 5\) and \(b = 152\). We compute \(|a - b|\):$|5 - 152| = 147$. \#\#\# Case 2: \(a - 4 = 149\)$a = 153$. Substituting into the equation for \(b\):$b = \frac{3(153) + 137}{153 - 4} = \frac{459 + 137}{149} = \frac{596}{149} = 4$. In this case, \(a = 153\) and \(b = 4\). Now compute \(|a - b|\): $|153 - 4| = 149$. Among the computed values of \(|a - b|\), we have: - From case 1: \(147\). - From case 2: \(149\). The minimal value of \(|a - b|\) is therefore: $\min(147, 149) = 147$. Finally, we write our answer in the required boxed format:$\boxed{147}$\\

\textit{Complete Response:}\#\#\# Review of Incorrect Explanation: The incorrect explanation makes several errors in the calculations and assumptions: 1. **Divisor Calculation**: It correctly identifies that \(3a + 137\) must be divisible by \(a + 4\) but fails to correctly analyze how to handle the conditions when \(a + 4\) equals the divisors of 125. The calculations for certain cases are also incorrect. 2. **Case Evaluations**: In Case 3 and Case 4, the evaluations of \(b\) based on the respective values of \(a\) are incorrect. The values of \(b\) derived from these calculations do not satisfy the original equation. 3. **Final Comparison**: The final step incorrectly states the minimum without correctly evaluating all possible valid cases. \#\#\# Corrected Explanation: To solve the equation \(ab - 3a + 4b = 137\) for positive integers \(a\) and \(b\), we will follow these steps: 1. **Rearranging the Equation**:    Start with the equation:   $   ab - 3a + 4b = 137   $.   Rearranging it gives:  $   ab + 4b - 3a = 137  $. 2. **Isolating \(b\)**:   We can express \(b\) in terms of \(a\):$   b(a + 4) = 3a + 137  $  Thus:  $   b = \frac{3a + 137}{a + 4}   $ For \(b\) to be an integer, \(3a + 137\) must be divisible by \(a + 4\). 3. **Finding Conditions for Divisibility**:   We perform polynomial long division of \(3a + 137\) by \(a + 4\):  - The leading term is \(3\). - Multiply \(3(a + 4) = 3a + 12\).  - Subtract:  $   (3a + 137) - (3a + 12) = 125 $Hence:   $  3a + 137 = 3(a + 4) + 125  $  This shows \(a + 4\) must be a divisor of \(125\). 4. **Finding Divisors of \(125\)**:  The positive divisors of \(125\) are \(1, 5, 25, 125\). We will set \(a + 4\) equal to each divisor and solve for \(a\) and \(b\):  - **Case 1**: \(a + 4 = 1 \Rightarrow a = -3\) (not valid) - **Case 2**: \(a + 4 = 5 \Rightarrow a = 1\)   $   b = \frac{3(1) + 137}{1 + 4} = \frac{3 + 137}{5} = \frac{140}{5} = 28   $   \(|1 - 28| = 27\) - **Case 3**: \(a + 4 = 25 \Rightarrow a = 21\)  $     b = \frac{3(21) + 137}{21 + 4} = \frac{63 + 137}{25} = \frac{200}{25} = 8     $     \(|21 - 8| = 13\)   - **Case 4**: \(a + 4 = 125 \Rightarrow a = 121\)     $     b = \frac{3(121) + 137}{121 + 4} = \frac{363 + 137}{125} = \frac{500}{125} = 4     $     \(|121 - 4| = 117\). 5. **Finding the Minimum**:   Now we have the values of \(|a - b|\) from the valid cases:   - From Case 2: \(27\)   - From Case 3: \(13\)   - From Case 4: \(117\)   The minimum value of \(|a - b|\) is:   $   \min(27, 13, 117) = 13   $. Thus, the minimal possible value of \(|a - b|\) is \(\boxed{13}\).
\end{tcolorbox}

\begin{tcolorbox}[sharpish corners, boxrule = 0pt, toprule = 4.5pt, enhanced, fuzzy shadow = {0pt}{-2pt}{-0.5pt}{0.5pt}{black!35}, title=Examples of complete responses from our LLM reflection framework]
\textbf{Example from GSM8K:}\\

\textit{Problem:} James creates a media empire.  He creates a movie for $2000.  Each DVD cost $6 to make.  He sells it for 2.5 times that much.  He sells 500 movies a day for 5 days a week.  How much profit does he make in 20 weeks?\\

\textit{Incorrect Response:} To solve the problem step-by-step, let's break down the costs, revenues, and profits associated with James's media empire. \#\#\# Step 1: Calculate the cost of producing one DVD. The cost to create one DVD is given as \$6. \#\#\# Step 2: Calculate the selling price of one DVD. James sells each DVD for 2.5 times the cost to make it. We calculate this as follows: $\text{Selling Price} = 2.5 \times \text{Cost}$, $\text{Selling Price} = 2.5 \times 6 = 15$. \#\#\# Step 3: Calculate the profit for one DVD. To find the profit for each DVD sold, we subtract the cost from the selling price: $\text{Profit per DVD} = \text{Selling Price} - \text{Cost}$, $\text{Profit per DVD} = 15 - 6 = 9$. \#\#\# Step 4: Calculate the total number of DVDs sold in one week. James sells 500 DVDs a day for 5 days a week. Thus, the total number of DVDs sold in one week is:$\text{Total DVDs sold in a week} = \text{DVDs sold per day} \times \text{Number of days}$, $\text{Total DVDs sold in a week} = 500 \times 5 = 2500$. \#\#\# Step 5: Calculate the total profit in one week. Now, we can calculate the total profit for one week by multiplying the profit per DVD by the total number of DVDs sold in a week: $\text{Total Profit per week} = \text{Profit per DVD} \times \text{Total DVDs sold in a week}$, $\text{Total Profit per week} = 9 \times 2500 = 22500$. \#\#\# Step 6: Calculate the total profit in 20 weeks. To find the total profit in 20 weeks, we multiply the weekly profit by 20:$\text{Total Profit in 20 weeks} = \text{Total Profit per week} \times 20$, $\text{Total Profit in 20 weeks} = 22500 \times 20 = 450000$. \#\#\# Final AnswerThe total profit James makes in 20 weeks is:$\boxed{450000}$\\

\textit{Complete Response:}\#\#\# Review of Incorrect Explanation: 1. **Initial Setup**: The explanation begins correctly by identifying the cost of producing one DVD as \$6, but it misses the initial cost of creating the movie, which is \$2000. This cost should be considered in the overall profit calculation.2. **Profit Calculation**: The profit per DVD calculation is accurate, yielding a profit of \$9 per DVD. However, the explanation neglects to take into account the one-time cost of creating the movie when calculating total profit over 20 weeks.3. **Weekly Sales Calculations**: The calculations for the total number of DVDs sold in one week are accurate, resulting in 2500 DVDs sold. However, the total profit calculation for a week does not account for the initial movie production cost, leading to an inflated total profit over the 20 weeks.4. **Final Calculation**: The final profit calculation incorrectly suggests that the profit over 20 weeks is simply 22500 multiplied by 20, without factoring in the \$2000 initial cost of the movie. \#\#\# Corrected Explanation: To solve the problem correctly, we must account for all costs and revenues, including the initial cost associated with creating the movie. \#\#\#\# Step 1: Calculate the cost of producing one DVD. The cost to create one DVD is given as: $\text{Cost per DVD} = 6 \text{ dollars}$. \#\#\#\# Step 2: Calculate the selling price of one DVD. James sells each DVD for 2.5 times its cost:$\text{Selling Price} = 2.5 \times \text{Cost} = 2.5 \times 6 = 15 \text{ dollars}$. \#\#\#\# Step 3: Calculate the profit for one DVD. The profit for each DVD sold is calculated as: $\text{Profit per DVD} = \text{Selling Price} - \text{Cost} = 15 - 6 = 9 \text{ dollars}$. \#\#\#\# Step 4: Calculate the total number of DVDs sold in one week. James sells 500 DVDs a day for 5 days a week: $\text{Total DVDs sold in a week} = 500 \times 5 = 2500$. \#\#\#\# Step 5: Calculate the total profit in one week. The total profit for one week can be calculated as: $\text{Total Profit per week} = \text{Profit per DVD} \times \text{Total DVDs sold in a week} = 9 \times 2500 = 22500 \text{ dollars}$. \#\#\#\# Step 6: Calculate the total profit in 20 weeks. To find the total profit over 20 weeks without considering the initial movie cost: $\text{Total Profit in 20 weeks} = \text{Total Profit per week} \times 20 = 22500 \times 20 = 450000 \text{ dollars}$. \#\#\#\# Step 7: Account for the initial cost of creating the movie. Since the initial cost to create the movie is \$2000, we need to subtract this from the total profit over 20 weeks: $\text{Net Profit} = \text{Total Profit in 20 weeks} - \text{Initial Cost} = 450000 - 2000 = 448000 \text{ dollars}$. \#\#\# Final Answer: The total profit James makes in 20 weeks, after accounting for the initial cost of creating the movie, is:$\boxed{448000}$
\end{tcolorbox}

\end{document}